המעבדה לבלשנות חישובית

Laboratory for Computational Linguistics

# Parsing with Typed Feature Structures

by

Shuly Wintner and Nissim Francez

הטכניון – מכון טכנולוגי לישראל, חיפה 32000 ישראל

Technion - Israel Institute of Technology, Haifa 32000 Israel

# Parsing with Typed Feature Structures*

Shuly Wintner Nissim Francez

Computer Science
Technion, Israel Institute of Technology
32000 Haifa, Israel
{shuly,francez}@cs.technion.ac.il

**Abstract**

In this paper we provide for parsing with respect to grammars expressed in a general TFS-based formalism, a restriction of ALE ([2]). Our motivation being the design of an abstract (WAM-like) machine for the formalism ([19]), we consider parsing as a computational process and use it as an operational semantics to guide the design of the control structures for the abstract machine.

We emphasize the notion of **abstract typed feature structures** (AFSs) that encode the essential information of TFSs and define unification over AFSs rather than over TFSs. We then introduce an explicit construct of **multi-rooted feature structures** (MRSs) that naturally extend TFSs and use them to represent phrasal signs as well as grammar rules. We also employ abstractions of MRSs and give the mathematical foundations needed for manipulating them. We formally define grammars and the languages they generate, and then describe a model for computation that corresponds to bottom-up chart parsing: grammars written in the TFS-based formalism are executed by the parser. We show that the computation is correct with respect to the independent definition. Finally, we discuss the class of grammars for which computations terminate and prove that termination can be guaranteed for off-line parsable grammars.

# 1 Introduction

Typed feature structures (TFSs) serve as a means for the specification of linguistic information in current linguistic formalisms such as HPSG ([14]) or Categorial Grammar ([8]). They are used for representing lexical items, phrases and rules. Usually, no mechanism for manipulating TFSs (e.g., parsing or generation algorithms) is inherent to the formalism. Current approaches to processing HPSG grammars either translate the grammar to Prolog (e.g., [2, 5, 6]) or specify it as a general constraint system ([21]).

In this paper we provide for parsing with respect to grammars expressed in a general TFS-based formalism, a restriction of ALE ([2]). Our motivation is the design of an abstract (WAM-like) machine for the formalism ([19]); we consider parsing as a computational process and use it as an operational semantics to guide the design of the control structures for the abstract machine. In this paper the machine is not discussed further.

Section 2 outlines the theory of TFSs of [1, 3]. We emphasize **abstract typed feature structures** (AFSs) that encode the essential information of TFSs and extend unification to AFSs. Section 3 introduces an explicit construct of **multi-rooted feature structures** (MRSs) that naturally extend TFSs, used to represent phrasal signs as well as grammar rules. Abstraction is extended to MRSs and the mathematical foundations needed for manipulating them is given. The concepts of grammars and the languages they generate are formally defined, and the TFS-based formalism is thus acquired a denotational semantics. In section 4 a model for computation, corresponding to bottom-up chart parsing for the formalism, is presented. The TFS-based formalism is

*A preliminary version of this paper appeared as [20].

thus endowed with an *operational* semantics. Next, we prove that both semantics coincide. Finally, we discuss the class of grammars for which computations terminate. We give a more relaxed definition for off-line parsability and prove that termination is guaranteed for off-line parsable grammars. The appendix contains a few examples of grammars and parsing.

The main contributions of this paper are:

- Formalization and explication of the notion of multi-rooted feature structures that are used implicitly in the computational linguistics literature;
- Concise definitions of a TFS-based linguistic formalism, based on abstract MRSs;
- Algebraic specification of a *parsing step* operator, $T_{G,w}$, that induces algebraic semantics for this formalism;
- Treatment of parsing as a model for computation, assigning operational semantics to the linguistic formalism;
- Specification and correctness proofs for parsing in this framework;
- A new definition for *off-line parsability*, less strict than the existing one, and termination proof for off-line parsable grammars.

# 2 Theory of Feature Structures

## 2.1 Feature Structures

The first part of this section summarizes some preliminary notions along the lines of [3]. For the following discussion we fix non-empty, finite, disjoint sets TYPES and FEATS of types and feature names, respectively. We assume that the set FEATS is totally ordered. We also fix an infinite set NODES of nodes, disjoint of TYPES and FEATS, each member of which is decorated by a type from TYPES through a fixed typing function $\theta$ : NODES $\rightarrow$ TYPES. The set NODES is 'rich' in the sense that for every $t \in$ TYPES, the set $\{q \in \text{NODES} \mid \theta(q) = t\}$ is infinite.

Below, the metavariable $T$ ranges over subsets of types, $t$ – over types, $f$ – over features and $q$ – over nodes. When dealing with partial functions the notation '$F(x)\downarrow$' means that $F$ is defined for the value $x$ and the symbol '$\uparrow$' means undefinedness. Whenever the result of an application of a partial function is used as an operand, it is meant that the function is defined for its arguments.

**Definition 2.1 (Type hierarchy)** *A partial order $\sqsubseteq$ over* TYPES $\times$ TYPES *is a* ***type hierarchy*** *(or* ***inheritance hierarchy****) if it is bounded complete, i.e., if every up-bounded subset $T$ of* TYPES *has a (unique) least upper bound, $\sqcup T$, referred to as the* ***unification*** *of the types in $T$.*

*If $t_1 \sqsubseteq t_2$ we say that $t_1$* ***subsumes****, or is* ***more general than****, $t_2$; $t_2$ is a* ***subtype*** *of (more* ***specific*** *than) $t_1$.*

*Let $\bot = \sqcup\phi$ be the most general type. Let the most specific type be $\top = \sqcup$*TYPES*. If $\sqcup T = \top$ we say that $T$ is* ***inconsistent****.*

**Definition 2.2 (Feature structures)** *A* ***feature structure*** *is a directed, connected, labeled graph consisting of a finite, nonempty set of nodes $Q \subseteq$* NODES*, a root $\bar{q} \in Q$, and a partial function $\delta : Q \times$* FEATS *$\rightarrow Q$ specifying the arcs such that every node $q \in Q$ is accessible from $\bar{q}$.*

The nodes of a feature structure are thus labeled by types while the arcs are labeled by feature names. The root $\bar{q}$ is a distinguished node from which all other nodes are reachable. A feature structure is of type $t$ when $\theta(\bar{q}) = t$. When we say that a feature structure $A$ *exists* we mean that no node of $A$ is typed $\top$.

We use upper-case letters (with or without tags, subscripts etc.) to refer to feature structures. We use $Q, \bar{q}, \delta$ (with the same tags or subscripts) to refer to constituents of feature structures.

Note that all feature structures are, by definition, graphs. Some grammatical formalisms used to have a special kind of feature structures, namely *atoms*; atoms are represented in our framework as nodes with no outgoing edges. For a discussion regarding the implications of such an approach, refer to [3, Chapter 8].

**Definition 2.3 (Paths)** *A **path** is a finite sequence of feature names, and the set* PATHS $=$ FEATS$^*$ *is the collection of paths. We use* $\pi, \alpha$ *(with or without subscripts) to refer to paths.* $\epsilon$ *is the empty path. The definition of* $\delta$ *is extended to paths in the natural way:*

$$\delta(q,\epsilon) = q$$
$$\delta(q, f\pi) = \delta(\delta(q,f),\pi)$$

*The paths of a feature structure* $A$ *are* $\Pi(A) = \{\pi \mid \pi \in \text{PATHS} \text{ and } \delta(\bar{q}_A, \pi)\downarrow\}$.

**Definition 2.4 (Cycles)** *A feature structure* $A = (Q, \bar{q}, \delta)$ *is **cyclic** if there exist a non-empty path* $\alpha \in$ PATHS *and a node* $q \in Q$ *such that* $\delta(q,\alpha) = q$. *It is **acyclic** otherwise.*

**Definition 2.5 (Reentrancy)** *A feature structure* $A$ *is **reentrant** iff there exist two different paths* $\pi_1, \pi_2$ *such that* $\delta(\bar{q}, \pi_1) = \delta(\bar{q}, \pi_2)$. *In this case the two paths are said to share the same value.*

## 2.2 Subsumption

**Definition 2.6 (Subsumption)** $A_1 = (Q_1, \bar{q}_1, \delta_1)$ ***subsumes*** $A_2 = (Q_2, \bar{q}_2, \delta_2)$ *(denoted by* $A_1 \sqsubseteq A_2$*) iff there exists a total function* $h : Q_1 \to Q_2$, *called a **subsumption morphism**, such that*

- $h(\bar{q}_1) = \bar{q}_2$
- *for every* $q \in Q_1$, $\theta(q) \sqsubseteq \theta(h(q))$
- *for every* $q \in Q_1$ *and for every* $f$ *such that* $\delta_1(q,f)\downarrow$, $h(\delta_1(q,f)) = \delta_2(h(q), f)$

$A_1 \sqsubset A_2$ *iff* $A_1 \sqsubseteq A_2$ *and* $A_1 \neq A_2$.

$h$ associates with every node in $Q_1$ a node in $Q_2$ with at least as specific a type; moreover, if an arc labeled $f$ connects $q$ with $q'$, then such an arc connects $h(q)$ with $h(q')$. If $A \sqsubseteq B$ then every path defined in $A$ is defined in $B$, and if two paths are reentrant in $A$ they are reentrant in $B$.

**Lemma 2.7** *If* $A \sqsubseteq B$ *then* $\Pi(A) \subseteq \Pi(B)$.

**Lemma 2.8** *If* $A \sqsubseteq B$ *then for every* $\pi_1, \pi_2 \in \Pi(A)$, *if* $\delta_A(\bar{q}_A, \pi_1) = \delta_A(\bar{q}_A, \pi_2)$ *then* $\delta_B(\bar{q}_B, \pi_1) = \delta_B(\bar{q}_B, \pi_2)$.

**Definition 2.9** *A partial order* $\succ$ *on* $D$ *is **well-founded** iff there exists no infinite decreasing sequence* $d_0 \succ d_1 \succ d_2 \succ \ldots$ *of elements of* $D$.

We prove below that subsumption of TFSs is well-founded iff they are acyclic.

**Lemma 2.10** *A TFS* $A$ *is acyclic iff* $\Pi(A)$ *is finite.*

**Proof:** If $A$ is cyclic, there exist a node $q \in Q$ and a non-empty path $\alpha$ that $\delta(q,\alpha) = q$. Let $\pi = \delta(\bar{q}, q)$, then the infinite set of paths $\{\pi\alpha^i \mid i \geq 0\}$ is contained in $\Pi(A)$. If $\Pi(A)$ is infinite then since $Q$ is finite, there exists a node $q \in Q$ that $\delta(q, \pi_i)\downarrow$ for an infinite number of different paths $\pi_i$. Since FEATS is finite, the out-degree of every node in $Q$ is finite; hence $q$ must be part of a cycle.

**Definition 2.11 (Rank)** *Let* $r : \text{TYPES} \to I\!N$ *be a total function such that* $r(t) < r(t')$ *if* $t \sqsubset t'$. *For an acyclic TFS* $A$, *let* $\Delta(A) = |\Pi(A)| - |Q_A|$ *and let* $\Theta(A) = \sum_{\pi \in \Pi(A)} r(\theta(\delta(\bar{q}, \pi)))$. *Define a rank for acyclic TFSs:* $rank(A) = \Delta(A) + \Theta(A)$.

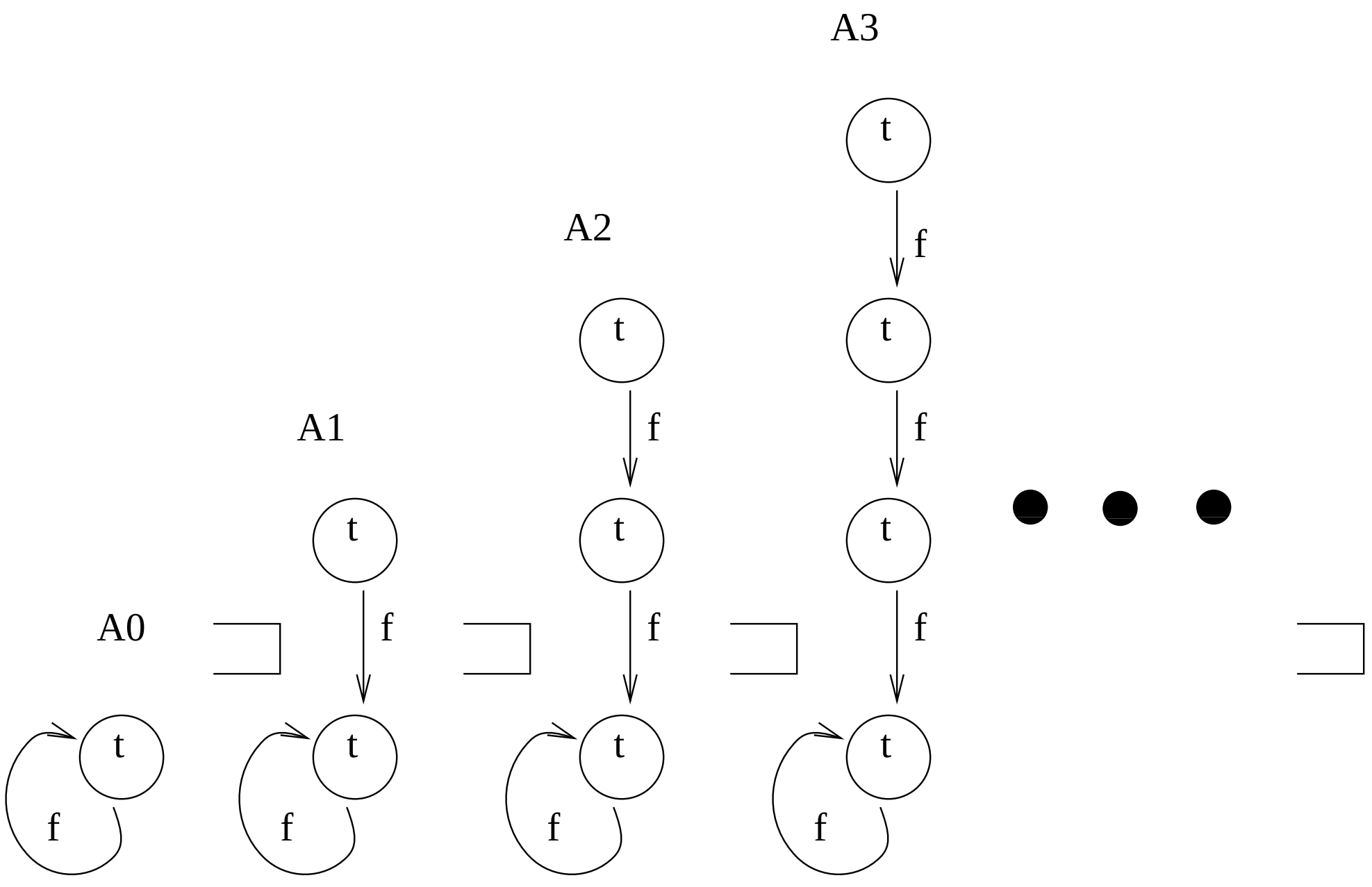

Figure 1: An infinite decreasing sequence of TFSs

By lemma 2.10, *rank* is well defined for acyclic TFSs. $\Delta(A)$ can be thought of as the number of reentrancies in $A$, or the number of different paths that lead to the same node in $A$. For every acyclic TFS $A$, $\Delta(A) \geq 0$ and hence $rank(A) \geq 0$.

**Lemma 2.12** *If* $A \sqsubset B$ *and both are acyclic then* $rank(A) < rank(B)$.

**Proof:** Assume $A \sqsubset B$ and both are acyclic; hence by lemma 2.7, $\Pi(A) \subseteq \Pi(B)$ and by lemma 2.10 both are finite. Let $h : Q_A \rightarrow Q_B$ be a subsumption morphism.

1. If $\Pi(A) = \Pi(B)$ then $|\Pi(A)| = |\Pi(B)|$. $A \sqsubset B$, hence either there exists a node $q \in Q_A$ that $\theta(q) \sqsubset \theta(h(q))$, and hence $\Theta(A) < \Theta(B)$ (while $\Delta(A) \leq \Delta(B)$); or (by lemma 2.8) there exist two paths $\pi_1, \pi_2$ that $\delta_A(\bar{q}_A, \pi_1) \neq \delta_A(\bar{q}_A, \pi_2)$, but $\delta_B(\bar{q}_B, \pi_1) = \delta_B(\bar{q}_B, \pi_2)$, in which case $\Delta(A) < \Delta(B)$ (while $\Theta(A) \leq \Theta(B)$). In any case, $rank(A) < rank(B)$.

2. If $\Pi(A) \subset \Pi(B)$ then $|\Pi(A)| < |\Pi(B)|$; as above, $\Theta(A) < \Theta(B)$. However, it might be the case that $|Q_A| < |Q_B|$. But for every node $q \in Q_B$ that is not the image of any node in $Q_A$, there exists a path $\pi$ such that $\delta(\bar{q}_B, \pi) = q$ and $\pi \notin \Pi(A)$. Hence $|\Pi(A)| - |Q_A| \leq |\Pi(B)| - |Q_B|$, and $rank(A) < rank(B)$.

**Theorem 2.13** *Subsumption of TFSs is not well-founded.*

**Proof:** Consider the infinite sequence of TFSs $A_0, A_1, \ldots$ depicted graphically in figure 1. For every $i \geq 0$, $A_i \sqsupset A_{i+1}$: to see that consider the morphism $h$ that maps $\bar{q}_{i+1}$ to $\bar{q}_i$ and $\delta_{i+1}(q, f)$ to $\delta_i(h(q), f)$ (i.e., the first $i+1$ nodes of $A_{i+1}$ are mapped to the first $i+1$ nodes of $A_i$, and the additional node of $A_{i+1}$ is mapped to the last node of $A_i$). Thus there exists a decreasing infinite sequence of cyclic TFSs and subsumption is not well-founded.

**Theorem 2.14** *Subsumption of acyclic TFSs is well-founded.*

**Proof:** For every acyclic TFS $A$, $rank(A)$ is finite and $rank(A) \geq 0$. By lemma 2.12, if $A \sqsubset B$ then $rank(A) < rank(B)$. If an infinite decreasing sequence of acyclic TFSs existed, *rank* would have mapped them to an infinite decreasing subsequence of $I\!N$, which is a contradiction. Hence subsumption is well-founded.

## 2.3 Abstract Feature Structures

The essential properties of a feature structure, excluding the identities of its nodes, can be captured by three components: the set of paths, the type that is assigned to every path, and the sets of paths that lead to the same node. In this section we elaborate on ideas presented in [12]; in contrast to the approach pursued in [3], we first define abstract feature structures and then show their relation to concrete ones. The representation of graphs as sets of paths is inspired by works on the semantics of concurrent programming languages, and the notion of fusion-closure is due to [4].

**Definition 2.15 (Alphabetic variants)** *Two feature structures $A$ and $B$ are* ***alphabetic variants*** *($A \sim B$) iff $A \sqsubseteq B$ and $B \sqsubseteq A$.*

Alphabetic variants have exactly the same structure, and corresponding nodes have the same types. Only the identities of the nodes distinguish them.

**Definition 2.16 (Abstract feature structures)** *A* ***pre- abstract feature structure*** *(pre-AFS) is a triple $\langle \Pi, \Theta, \approx \rangle$, where*

- $\Pi \subseteq$ PATHS *is a non-empty set of paths*
- $\Theta : \Pi \rightarrow$ TYPES *is a total function, assigning a type to every path*
- $\approx \subseteq \Pi \times \Pi$ *is a relation specifying reentrancy*

*An* ***abstract feature structure*** *(AFS) is a pre-AFS $A$ for which the following requirements hold:*

- *$\Pi$ is prefix-closed: if $\pi\alpha \in \Pi$ then $\pi \in \Pi$ (where $\pi, \alpha \in$ PATHS)*
- *$A$ is fusion-closed: if $\pi\alpha \in \Pi$ and $\pi'\alpha' \in \Pi$ and $\pi \approx \pi'$ then $\pi\alpha' \in \Pi$ (as well as $\pi'\alpha \in \Pi$) and $\pi\alpha' \approx \pi'\alpha'$ (as well as $\pi'\alpha \approx \pi\alpha$)*
- *$\approx$ is an equivalence relation with a finite index (with $[\approx]$ the set of its equivalence classes)*
- *$\Theta$ respects the equivalence: if $\pi_1 \approx \pi_2$ then $\Theta(\pi_1) = \Theta(\pi_2)$*

Abstract features structures can be related to concrete ones in a natural way: If $A = (Q, \bar{q}, \delta)$ is a TFS then $Abs(A) = \langle \Pi_A, \Theta_A, \approx_A \rangle$ is defined by:

- $\Pi_A = \{\pi \mid \delta(\bar{q}, \pi)\downarrow\}$
- $\Theta_A(\pi) = \theta(\delta(\bar{q}, \pi))$
- $\pi_1 \approx_A \pi_2$ iff $\delta(\bar{q}, \pi_1) = \delta(\bar{q}, \pi_2)$

**Lemma 2.17** *If $A$ is a feature structure then $Abs(A)$ is an abstract feature structure.*

**Proof:**

1. $\Pi$ is prefix-closed: $\Pi = \{\pi \mid \delta(\bar{q}, \pi)\downarrow\}$. If $\pi\alpha \in \Pi$ then $\delta(\bar{q}, \pi\alpha)\downarrow$ and by the definition of $\delta$, $\delta(\bar{q}, \pi)\downarrow$, too.
2. $Abs(A)$ is fusion-closed: Suppose that $\pi\alpha \in \Pi, \pi'\alpha' \in \Pi$ and $\pi \approx \pi'$. Then $\delta(\bar{q}, \pi) = \delta(\bar{q}, \pi')$. Hence $\delta(\bar{q}, \pi\alpha')\downarrow$ (therefore $\pi\alpha' \in \Pi$), and $\delta(\bar{q}, \pi\alpha') = \delta(\pi'\alpha')$, therefore $\pi\alpha' \approx \pi'\alpha'$. In the same way, $\pi'\alpha \in \Pi$ and $\pi'\alpha \approx \pi'\alpha'$.
3. $\approx$ is an equivalence relation with a finite index: $\pi_1 \approx \pi_2$ iff $\delta(\bar{q}, \pi_1) = \delta(\bar{q}, \pi_2)$, namely iff $\pi_1$ and $\pi_2$ lead to the same node (from $\bar{q}$) in $A$. Hence $\approx$ is an equivalence relation and since $Q$ is finite, $\approx$ has a finite index.
4. $\Theta$ respects the equivalence: $\Theta(\pi) = \theta(\delta(\bar{q}, \pi))$ and if $\pi_1 \approx \pi_2$ then $\delta(\bar{q}, \pi_1) = \delta(\bar{q}, \pi_2)$, hence $\Theta(\pi_1) = \Theta(\pi_2)$.

For the reverse direction, consider an AFS $A = \langle \Pi, \Theta, \approx \rangle$. First construct a 'pseudo-TFS', $Conc(A) = (Q, \bar{q}, \delta)$, that differs from a TFS only in that its nodes are not drawn from the set Nodes. Let $Q = \{q_{[\pi]} \mid [\pi] \in [\approx]\}$, making use of the fact that '$\approx$' is of finite index. Let $\theta(q_{[\pi]}) = \Theta(\pi)$ for every node – since $A$ is an AFS, $\Theta$ respects the equivalence and therefore $\theta$ is representative-independent. Let $\bar{q} = q_{[\epsilon]}$ and $\delta(q_{[\pi]}, f) = q_{[\pi f]}$ for every node $q_{[\pi]}$ and feature $f$. Since $A$ is fusion-closed, $\delta$ is representative-independent. By injecting $Q$ into Nodes, making use of the richness of Nodes, a concrete TFS $Conc(A)$ is obtained, representing the equivalence class of alphabetic variants that can be obtained that way. We abuse the notation $Conc(A)$ in the sequel to refer to this set of alphabetic variants. Figure 2 depicts an example feature structure, represented both as an AVM and as a graph, along with its abstraction.

Graph representation:

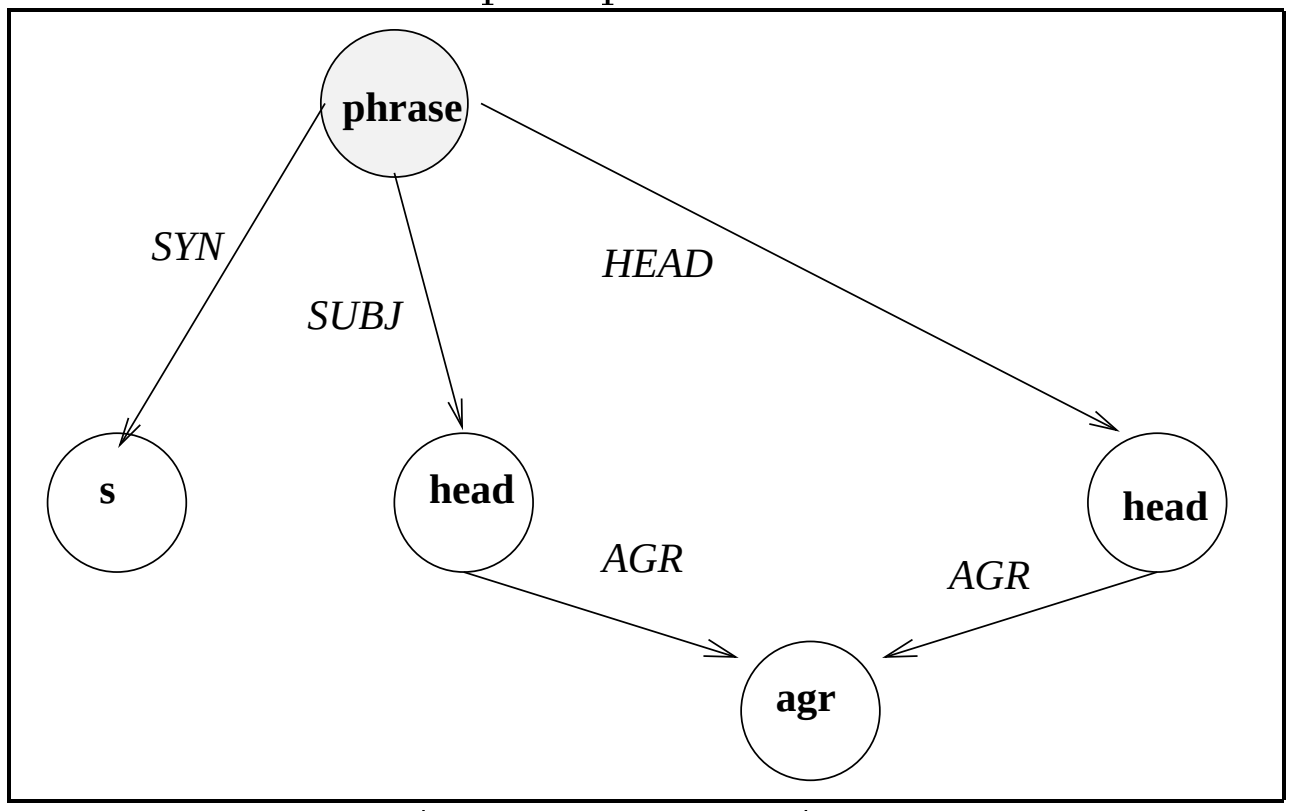

AVM representation:

$$\begin{bmatrix} \textbf{phrase} & \\ SYN: & [\textbf{s}] \\ SUBJ: & \begin{bmatrix} \textbf{head} \\ AGR: \boxed{3}[\textbf{agr}] \end{bmatrix} \\ HEAD: & \begin{bmatrix} \textbf{head} \\ AGR: \boxed{3} \end{bmatrix} \end{bmatrix}$$

Abstract representation:

$$\Pi = \{\epsilon, \text{SYN}, \text{SUBJ}, \text{SUBJ AGR}, \text{HEAD}, \text{HEAD AGR}\}$$

| $\approx$ | $\epsilon$ | SYN | SUBJ | SUBJ AGR | HEAD | HEAD AGR | $\Theta$ |
|---|---|---|---|---|---|---|---|
| $\epsilon$ | + | | | | | | **phrase** |
| SYN | | + | | | | | **s** |
| SUBJ | | | + | | | | **head** |
| SUBJ AGR | | | | + | | + | **agr** |
| HEAD | | | | | + | | **head** |
| HEAD AGR | | | | + | | + | **agr** |

Figure 2: A feature structure

**Theorem 2.18** *If $A' \in Conc(A)$ then $Abs(A') = A$.*

**Proof:** Let $A = \langle \Pi_A, \Theta_A, \approx_A \rangle, A' = (Q, \bar{q}, \delta), Abs(A') = \langle \Pi, \Theta, \approx \rangle$. If $A' \in Conc(A)$ then $Q$ can be mapped by a one-to-one function to the set of equivalence classes of $\approx_A$ and $\delta$ determines the paths in $\Pi_A$. By the definition of $Abs, \Pi = \Pi_A$. Given a path $\pi \in \Pi_A, \Theta(\pi) = \theta(\delta(\bar{q}, \pi)) = \theta(q_{[\pi]}) = \Theta_A(\pi)$. If $\pi_1 \approx_A \pi_2$ then $\delta(\bar{q}, \pi_1) = \delta(\bar{q}, \pi_2)$ (since $A$ is fusion-closed) and hence $\pi_1 \approx \pi_2$.

AFSs can be partially ordered: $\langle \Pi_A, \Theta_A, \approx_A \rangle \preceq \langle \Pi_B, \Theta_B, \approx_B \rangle$ iff $\Pi_A \subseteq \Pi_B, \approx_A \subseteq \approx_B$ and for every $\pi \in \Pi_A, \Theta_A(\pi) \sqsubseteq \Theta_B(\pi)$. This order corresponds to the subsumption ordering on TFSs, as the following theorems show.

**Theorem 2.19** *$A \sqsubseteq B$ iff $Abs(A) \preceq Abs(B)$.*

**Proof:** Let $Abs(A) = \langle \Pi_A, \Theta_A, \approx_A \rangle, Abs(B) = \langle \Pi_B, \Theta_B, \approx_B \rangle$. Assume that $A \sqsubseteq B$, that is, a subsumption morphism $h : Q_A \to Q_B$ exists. If $\pi \in \Pi_A$ then (from the definition of $Abs(A)$) $\delta_A(\bar{q}_A, \pi)\downarrow$, that is, there exists a sequence $q_0, q_1, \ldots, q_n$ of nodes and a sequence $f_1, \ldots, f_n$ of features such that for every $i$, $0 \leq i < n, \delta_A(q_i, f_{i+1}) = q_{i+1}$, $q_0 = \bar{q}_A$ and $\pi = f_1 \cdots f_n$. Due to the subsumption morphism, there exists a sequence of nodes $h(q_0), \ldots, h(q_n)$ such that $\delta_B(h(q_i), f_{i+1}) = h(q_{i+1})$ for every $i$, $0 \leq i < n$, and $h(q_0) = \bar{q}_B$. Hence $\pi \in \Pi_B$. Moreover, since $A \sqsubseteq B$, for every node $q$, $\theta(q) \sqsubseteq \theta(h(q))$. In particular, $\theta(q_n) \sqsubseteq \theta(h(q_n))$ and thus $\Theta_A(\pi) \sqsubseteq \Theta_B(\pi)$. Now suppose that two paths $\pi_1, \pi_2$ are reentrant in $A$. By the definition of subsumption, $\pi_1$ and $\pi_2$ are reentrant in $B$, too. Therefore $\approx_A \subseteq \approx_B$.
If $Abs(A) \preceq Abs(B)$, construct a function $h : Q_A \to Q_B$ such that $h(\bar{q}_A) = \bar{q}_B$ and for every $q \in Q_A, h(\delta_A(q, f)) = \delta_B(h(q), f)$. Trivially, $h$ is total and $h(\bar{q}_A) = \bar{q}_B$. Also, if $\delta_A(q, f)\downarrow$ then $h(\delta_A(q, f)) = \delta_B(h(q), f)$. To show that $\theta(q) \sqsubseteq \theta(h(q))$ for every $q$, consider a path $\pi$ leading from $\bar{q}_A$ to $q$. Since $Abs(A) \preceq Abs(B), \Theta_A(\pi) \sqsubseteq \Theta_B(\pi)$ and hence $\theta(q) \sqsubseteq \theta(h(q))$. Hence $h$ is a subsumption morphism.

**Theorem 2.20** *For every $A \in Conc(A'), B \in Conc(B'), A \sqsubseteq B$ iff $A' \preceq B'$.*

**Proof:** Select some $A \in Conc(A'), B \in Conc(B')$. If $A \sqsubseteq B$ then, by theorem 2.19, $Abs(A) \preceq Abs(B)$. By the definition of $Conc$, $Abs(A) = A'$ and $Abs(B) = B'$, so that $A' \preceq B'$.
If $A' \preceq B'$, construct a function $h : Q_A \to Q_B$ as follows: First, let $h(\bar{q}_A) = \bar{q}_B$. Then, perform a depth-first search on the graph $A$ and for every node $q' = \delta_A(q, f)$ encountered, if $h(q')\uparrow$ set $h(q') = \delta_B(h(q), f)$. The order of the search is irrelevant: since $A' \preceq B', \approx_{A'} \subseteq \approx_{B'}$ and therefore if $\pi_1 \approx_{A'} \pi_2$ then $\pi_1 \approx_{B'} \pi_2$. Since $A' \preceq B', \Pi_{A'} \subseteq \Pi_{B'}$ and hence $\delta_B(h(q), f)$ is defined whenever $\delta_A(q, f)$ is defined. Hence $h$ is total and $h(\bar{q}_A) = \bar{q}_B$. For every node $q \in Q_A$, some path $\pi$ exists that leads from $\bar{q}_A$ to $q$ and from $\bar{q}_B$ to $h(q)$. $\Theta_A(\pi) \sqsubseteq \Theta_B(\pi)$, and therefore $\theta(q) \sqsubseteq \theta(h(q))$. Hence $h$ is a subsumption morphism.

**Corollary 2.21** *$A \sim B$ iff $Abs(A) = Abs(B)$.*

**Proof:** Immediate from theorem 2.19.

**Corollary 2.22** *$Conc(A) \sim Conc(B)$ iff $A = B$.*

**Proof:** Immediate from theorem 2.20.

## 2.4 Unification

As there exists a one to one correspondence between abstract feature structures and (alphabetic variants of) concrete ones, we define unification directly over AFSs. This leads to a simpler definition that captures the essence of the operation better than the traditional definition. We use the term 'unification' to refer to both the operation and its result.

**Lemma 2.23** *If $A = \langle \Pi_A, \Theta_A, \approx_A \rangle$ is a pre-AFS then there exists a pre-AFS $B = \langle \Pi_B, \Theta_B, \approx_B \rangle$ such that $B$ is the least extension of $A$ to a fusion-closed structure and $\Theta_B(\pi) = \Theta_A(\pi)$ for every $\pi \in \Pi_A$.*

**Lemma 2.24** *If $A = \langle \Pi_A, \Theta_A, \approx_A \rangle$ is a pre-AFS then there exists a pre-AFS $B = \langle \Pi_B, \Theta_B, \approx_B \rangle$ such that $\Pi_A = \Pi_B, \Theta_A = \Theta_B$ and $\approx_B$ is the least extension of $\approx_A$ to an equivalence relation.*

**Definition 2.25 (Closure operations)** *Let $Cl$ be a fusion-closure operation on pre-AFSs: $Cl(A) = A'$, where $A'$ is the least extension of $A$ to a fusion-closed structure. Let $Eq(\langle \Pi, \Theta, \approx \rangle) = \langle \Pi, \Theta, \approx' \rangle)$ where $\approx'$ is the least extension of $\approx$ to an equivalence relation. Let $Ty(\langle \Pi, \Theta, \approx \rangle) = \langle \Pi, \Theta', \approx \rangle$ where $\Theta'(\pi) = \bigsqcup_{\pi' \approx \pi} \Theta(\pi)$.*

**Definition 2.26 (Unification)** *The unification $A \sqcup B$ of two AFSs $A = \langle \Pi_A, \Theta_A, \approx_A \rangle$ and $B = \langle \Pi_B, \Theta_B, \approx_B \rangle$ is the AFS $C' = Ty(Eq(Cl(C)))$, where:*

- $C = \langle \Pi_C, \Theta_C, \approx_C \rangle$
- $\Pi_C = \Pi_A \cup \Pi_B$
- $\Theta_C(\pi) = \begin{cases} \Theta_A(\pi) \sqcup \Theta_B(\pi) & \text{if } \pi \in \Pi_A \text{ and } \pi \in \Pi_B \\ \Theta_A(\pi) & \text{if } \pi \in \Pi_A \text{ only} \\ \Theta_B(\pi) & \text{if } \pi \in \Pi_B \text{ only} \end{cases}$
- $\approx_C = \approx_A \cup \approx_B$

*The unification* ***fails*** *if there exists a path $\pi \in \Pi_{C'}$ such that $\Theta_{C'}(\pi) = \top$.*

**Lemma 2.27** *$Cl$ preserves prefixes: If $A$ is a prefix-closed pre-AFS and $A' = Cl(A)$ then $A'$ is prefix-closed.*

**Proof:** Let $\pi$ be a path in $\Pi'$. If $\pi \in \Pi$ then every prefix of $\pi$ is in $\Pi'$, since $\Pi$ is prefix-closed and $Cl$ only adds paths. Suppose that $\pi \in \Pi' \setminus \Pi$. Then there exist $\pi_1, \pi_2, \alpha_1, \alpha_2 \in$ PATHS such that $\pi_1\alpha_1 \in \Pi$ and $\pi_2\alpha_2 \in \Pi$ and $\pi_1 \approx \pi_2$ and $\pi = \pi_1\alpha_2$ (otherwise, $\pi$ can be removed from $\Pi'$, in contradiction to the minimality of $Cl$). If $\pi'$ is a prefix of $\pi$ than either $\pi'$ is a prefix of $\pi_1$, in which case $\pi' \in \Pi$ since $\Pi$ is prefix-closed, or $\pi' = \pi_1\alpha'$ for some $\alpha'$ that is a prefix of $\alpha$. Since $\Pi$ is prefix-closed, $\pi_1\alpha' \in \Pi$ and $\pi_2\alpha' \in \Pi$. Therefore, as $\pi_1 \approx \pi_2, \pi_1\alpha'$ is added to $\Pi'$ by the closure operation.

**Lemma 2.28** *$Eq$ preserves prefixes and fusions: If $A$ is a prefix- and fusion-closed pre-AFS and $A' = Eq(A)$ then $A'$ is prefix- and fusion-closed.*

**Proof:** $Eq$ extends $\approx$ to an equivalence relation. Since only $\approx$ is modified, prefix-closure is trivially maintained. Select a pair $(\pi_1, \pi_2) \in \approx' \setminus \approx$. Then either (1) $\pi_2 = \pi_1$; (2) $\pi_2 \approx \pi_1$; or (3) there exists a path $\pi_3$ such that $\pi_1 \approx \pi_3$ and $\pi_3 \approx \pi_2$. Trivially, (1) and (2) preserve the closure properties. In the case of (3), to show that fusion-closure is maintained we have to show that if $\pi_1\alpha_1 \in \Pi'$ and $\pi_2\alpha_2 \in \Pi'$ then $\pi_1\alpha_2 \in \Pi'$ and $\pi_1\alpha_2 \approx' \pi_2\alpha_2$. Since $\Pi = \Pi', \pi_1\alpha_1 \in \Pi$ and $\pi_2\alpha_2 \in \Pi$. Since $\Pi$ is fusion-closed and $\pi_2 \approx \pi_3, \pi_3\alpha_2 \in \Pi$ and $\pi_3\alpha_2 \approx \pi_2\alpha_2$. Since $\pi_1 \approx \pi_3, \pi_1\alpha_2 \in \Pi$ and $\pi_1\alpha_2 \approx \pi_3\alpha_2$, too. $\approx'$ is an extension of $\approx$ to an equivalence relation, and thus $\pi_1\alpha_2 \approx' \pi_2\alpha_2$.

**Corollary 2.29** *If $A$ and $B$ are AFSs, then so is $A \sqcup B$.*

**Proof:** If $A$ and $B$ are AFSs then the pre-AFS $C$, defined as in 2.26, is prefix-closed (since $A$ and $B$ are). $Cl(C)$ is prefix- and fusion-closed, as is $Eq(Cl(C))$ in which, additionally, $\approx$ is an equivalence relation. $Ty(Eq(Cl(C)))$ is an AFS, since $Ty$ only modifies $\Theta$ such that it respects the equivalences.

$C'$ is the smallest AFS that contains $\Pi_C$ and $\approx_C$. Since $\Pi_A$ and $\Pi_B$ are prefix-closed, so is $\Pi_C$. However, $\Pi_C$ and $\approx_C$ might not be fusion-closed. This is why $Cl$ is applied to them. As a result of its application, new paths and equivalence classes might be added. By lemma 2.27, if a path is added all its prefixes are added, too, so the prefix-closure is preserved. Then, $Eq$ extends $\approx$ to an equivalence relation, without harming the prefix- and fusion-closure properties (by lemma 2.28). Finally, $Ty$ sees to it that $\Theta$ respects the equivalences.

**Lemma 2.30** *Unification is commutative: $A \sqcup B = B \sqcup A$.*

**Proof:** Observe that unification is defined using set union ($\cup$) and type unification ($\sqcup$) which are commutative. Therefore, the unification is commutative, too.

**Lemma 2.31** *Unification is associative: $(A \sqcup B) \sqcup C = A \sqcup (B \sqcup C)$.*

**Proof:** as above.

The result of a unification can differ from any of its arguments in three ways: paths that were not present can be added; the types of nodes can become more specific; and reentrancies can be added, that is, the number of equivalence classes of paths can decrease. Consequently, the result of a unification is always more specific than any of its arguments.

**Theorem 2.32** *If* $C' = A \sqcup B$ *then* $A \preceq C'$.

**Proof:** $\Pi_C = \Pi_A \cup \Pi_B$ and hence $\Pi_A \subseteq \Pi_C$. $\approx_C = \approx_A \cup \approx_B$ and hence $\approx_A \subseteq \approx_C$. If $\pi \in \Pi_A$ then $\Theta_C(\pi) = \Theta_A(\pi)$ or $\Theta_C(\pi) = \Theta_A(\pi) \sqcup \Theta_B(\pi)$, and in any case $\Theta_A(\pi) \sqsubseteq \Theta_C(\pi)$. $Cl$ and $Eq$ cannot remove paths or equivalences and $Ty$ only makes types more specific, and therefore $A \preceq C'$.

**Theorem 2.33** $A \sqcup B = A$ *iff* $B \preceq A$.

**Proof:** Suppose $B \preceq A$. Then $\Pi_B \subseteq \Pi_A, \approx_B \subseteq \approx_A$ and for every $\pi \in \Pi_B, \Theta_B(\pi) \sqsubseteq \Theta_A(\pi)$. $A \sqcup B = Ty(Eq(Cl(C)))$ where $C = \langle \Pi_C, \Theta_C, \approx_C \rangle$ and

- $\Pi_C = \Pi_A \cup \Pi_B = \Pi_A$
- $\Theta_C(\pi) = \begin{cases} \Theta_A(\pi) \sqcup \Theta_B(\pi) & \text{if } \pi \in \Pi_A \text{ and } \pi \in \Pi_B \\ \Theta_A(\pi) & \text{if } \pi \in \Pi_A \text{ only} \\ \Theta_B(\pi) & \text{if } \pi \in \Pi_B \text{ only} \end{cases} = \Theta_A(\pi)$
- $\approx_C = \approx_A \cup \approx_B = \approx_A$

Hence $A = C$ and therefore $A \sqcup B = A$.
Suppose $A \sqcup B = A$ and assume toward a contradiction that $B \not\preceq A$. Then at least one of the following cases holds:

- $\Pi_B \not\subseteq \Pi_A$. Then there exists $\pi \in \Pi_B \cup \Pi_A$ that $\pi \notin \Pi_A$ and hence $A \sqcup B \neq A$.
- There exists some $\pi$ such that $\Theta_B(\pi) \not\sqsubseteq \Theta_A(\pi)$. Then $\Theta_A(\pi) \sqcup \Theta_B(\pi) \neq \Theta_A(\pi)$ and hence $A \sqcup B \neq A$.
- $\approx_B \not\subseteq \approx_A$. Then there exist $\pi_1, \pi_2$ such that $(\pi_1 \approx_B \pi_2)$ but not $(\pi_1 \approx_A \pi_2)$. Hence $(\approx_A \cup \approx_B) \neq \approx_A$ and $A \sqcup B \neq A$.

TFSs (and therefore AFSs) can be seen as a generalization of first-order terms (FOTs) (see [1]). Accordingly, AFS unification resembles FOT unification; however, the notion of *substitution* that is central to the definition of FOT unification is missing here, and as far as we know, no analog to substitutions in the domain of feature structures was ever presented.

# 3 Multi-rooted Structures

To be able to represent complex linguistic information, such as phrase structure, the notion of feature structures has to be extended. HPSG does so by introducing special features, such as DTRS (daughters), to encode trees in TFSs. This solution requires a declaration of the special features, along with their intended meaning; such a declaration is missing in [14]. An alternative technique is employed by Shieber ([16]): a denumerable set of special features, namely $0, 1, \ldots$, are added to encode the order of daughters in a tree. In a typed system such as ours, this method would necessitate the addition of special types as well; in general, no bound can be placed on the number of features and types necessary to state rules.

As a more coherent, mathematically elegant solution, we define the notion of multi-rooted structures that naturally extend TFSs. These structures provide a means to represent phrasal signs and grammar rules. They are used implicitly in the computational linguistics literature, but to the best of our knowledge no explicit, formal theory of these structures and their properties was formulated before.

**Definition 3.1 (Multi-rooted structures)** *A **multi-rooted feature structure** (MRS) is a pair $\langle \bar{Q}, G\rangle$ where $G = \langle Q, \delta\rangle$ is a finite, directed, labeled graph consisting of a set $Q \subseteq$ NODES of nodes and a partial function $\delta : Q \times$ FEATS $\rightarrow Q$ specifying the arcs, and where $\bar{Q}$ is an ordered, (repetition-free) list of distinguished nodes in $Q$ called **roots**. $G$ is not necessarily connected, but the union of all the nodes reachable from all the roots in $\bar{Q}$ is required to yield exactly $Q$. The **length** of a MRS is the number of its roots, $|\bar{Q}|$. $\lambda$ denotes the empty MRS, where $Q = \phi$.*

Meta-variables $\sigma, \rho$ range over MRSs, and $\delta, Q$ and $\bar{Q}$ over their constituents. If $\langle \bar{Q}, G\rangle$ is a MRS and $\bar{q}_i$ is a root in $\bar{Q}$ then $\bar{q}_i$ naturally induces a feature structure $Pr(\bar{Q}, i) = (Q_i, \bar{q}_i, \delta_i)$, where $Q_i$ is the set of nodes reachable from $\bar{q}_i$ and $\delta_i = \delta|_{Q_i}$.

One can view a MRS $\langle \bar{Q}, G\rangle$ as an ordered sequence $\langle A_1, \ldots, A_n\rangle$ of (not necessarily disjoint) feature structures, where $A_i = Pr(\bar{Q}, i)$ for $1 \leq i \leq n$. Note that such an ordered list of feature structures is not a sequence in the mathematical sense: removing an element from the list may effect the other elements (due to reentrancy among elements). Nevertheless, we can think of a MRS as a sequence where a subsequence is obtained by taking a subsequence of the roots and considering only the feature structures they induce. We use the two views interchangeably. Figure 3 depicts a MRS and its view as a sequence of feature structures. The shaded nodes (ordered from left to right) constitute $\bar{Q}$.

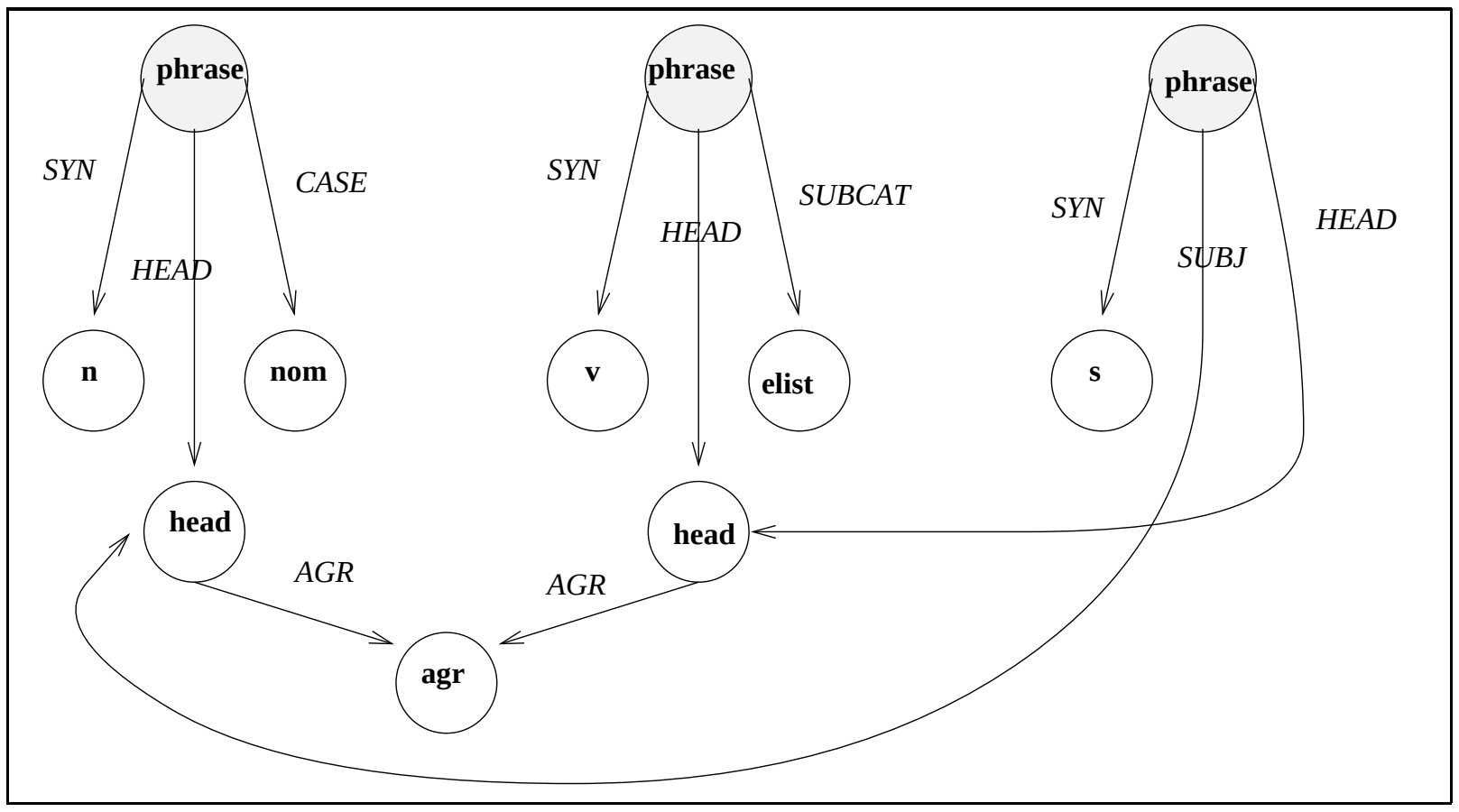

$$\begin{bmatrix}\textbf{phrase} & \\ SYN: & [\textbf{n}]\\ HEAD: & \boxed{1}\begin{bmatrix}\textbf{head} & \\ AGR: & \boxed{3}[\textbf{agr}]\end{bmatrix}\\ CASE: & [\textbf{nom}]\end{bmatrix}\begin{bmatrix}\textbf{phrase} & \\ SYN: & [\textbf{v}]\\ HEAD: & \boxed{2}\begin{bmatrix}\textbf{head} & \\ AGR: & \boxed{3}\end{bmatrix}\\ SUBCAT: & [\textbf{elist}]\end{bmatrix}\begin{bmatrix}\textbf{phrase} & \\ SYN: & [\textbf{s}]\\ SUBJ: & \boxed{1}\\ HEAD: & \boxed{2}\end{bmatrix}$$

Figure 3: A graph- and AVM- representation of a MRS

Subsumption is extended to MRSs as follows:

**Definition 3.2 (Subsumption of multi-rooted structures)** *A MRS $\sigma = \langle \bar{Q}, G\rangle$ **subsumes** a MRS $\sigma' = \langle \bar{Q}', G'\rangle$ (denoted by $\sigma \sqsubseteq \sigma'$) if $|\bar{Q}| = |\bar{Q}'|$ and there exists a total function $h : Q \rightarrow Q'$ such that:*

- *for every root $\bar{q}_i \in \bar{Q}, h(\bar{q}_i) = \bar{q}'_i$*
- *for every $q \in Q$, $\theta(q) \sqsubseteq \theta'(h(q))$*
- *for every $q \in Q$ and $f \in$ FEATS, if $\delta(q, f){\downarrow}$ then $h(\delta(q, f)) = \delta'(h(q), f)$*

We define abstract multi-rooted structures in an analog way to abstract feature structures.

**Definition 3.3 (Abstract multi-rooted structures)** *A **pre- abstract multi rooted structure** (pre-AMRS) is a quadruple* $A = \langle Ind, \Pi, \Theta, \approx \rangle$*, where:*

- $Ind$*, the **indices** of* $A$*, is the set* $\{1, \ldots, n\}$ *for some* $n$
- $\Pi \subseteq Ind \times \textsc{Paths}$ *is a set of indexed paths, such that for each* $i \in Ind$ *there exists some* $\pi \in \textsc{Paths}$ *that* $(i, \pi) \in \Pi$.
- $\Theta : \Pi \rightarrow \textsc{Types}$ *is a total type-assignment function*
- $\approx \subseteq \Pi \times \Pi$ *is a relation*

*An **abstract multi-rooted structure** (AMRS) is a pre-AMRS* $A$ *for which the following requirements, naturally extending those of AFSs, hold:*

- $\Pi$ *is prefix-closed: if* $(i, \pi\alpha) \in \Pi$ *then* $(i, \pi) \in \Pi$
- $A$ *is fusion-closed: if* $(i, \pi\alpha) \in \Pi$ *and* $(i', \pi'\alpha') \in \Pi$ *and* $(i, \pi) \approx (i', \pi')$ *then* $(i, \pi\alpha') \in \Pi$ *(as well as* $(i', \pi'\alpha) \in \Pi$*), and* $(i, \pi\alpha') \approx (i', \pi'\alpha')$ *(as well as* $(i', \pi'\alpha) \approx (i, \pi\alpha)$*)*
- $\approx$ *is an equivalence relation with a finite index*
- $\Theta$ *respects the equivalence: if* $(i_1, \pi_1) \approx (i_2, \pi_2)$ *then* $\Theta(i_1, \pi_1) = \Theta(i_2, \pi_2)$

The **length** of an AMRS $A$ is $len(A) = |Ind_A|$. We use $\lambda$ to denote the empty AMRS, too, where $Ind_\lambda = \phi$ and $\Pi_\lambda = \phi$ (so that $len(\lambda) = 0$).

The closure operations $Cl$ and $Eq$ are naturally extended to AMRSs: If $A$ is a pre-AMRS then $Cl(A)$ is the least extension of $A$ that is prefix- and fusion-closed, and $Eq(A)$ is the least extension of $A$ to a pre-AMRS in which $\approx$ is an equivalence relation. In addition, $Ty(\langle Ind, \Pi, \Theta, \approx \rangle) = \langle Ind, \Pi, \Theta', \approx \rangle$ where $\Theta'(i, \pi) = \bigsqcup_{(i', \pi') \approx (i, \pi)} \Theta(i', \pi')$. The partial order $\preceq$ is extended to AMRSs: $\langle Ind_A, \Pi_A, \Theta_A, \approx_A \rangle \preceq \langle Ind_B, \Pi_B, \Theta_B, \approx_B \rangle$ iff $Ind_A = Ind_B, \Pi_A \subseteq \Pi_B, \approx_A \subseteq \approx_B$ and for every $(i, \pi) \in \Pi_A, \Theta_A(i, \pi) \sqsubseteq \Theta_B(i, \pi)$. In the rest of this paper we overload the symbol '$\sqsubseteq$' so that it denotes subsumption of AMRSs as well as MRSs.

AMRSs, too, can be related to concrete ones in a natural way: If $\sigma = \langle \bar{Q}, G \rangle$ is a MRS then $Abs(\sigma) = \langle Ind_\sigma, \Pi_\sigma, \Theta_\sigma, \approx_\sigma \rangle$ is defined by:

- $Ind_\sigma = \{1, \ldots, |\bar{Q}|\}$
- $\Pi_\sigma = \{(i, \pi) \mid \delta(\bar{q}_i, \pi)\downarrow\}$
- $\Theta_\sigma(i, \pi) = \theta(\delta(\bar{q}_i, \pi))$
- $(i, \pi_1) \approx_\sigma (j, \pi_2)$ iff $\delta(\bar{q}_i, \pi_1) = \delta(\bar{q}_j, \pi_2)$

It is easy to see that $Abs(\sigma)$ is an AMRS. In particular, notice that for every $i \in Ind_\sigma$ there exists a path $\pi$ such that $(i, \pi) \in \Pi_\sigma$ since for every $i, \delta(\bar{q}_i, \epsilon)\downarrow$. The reverse operation, $Conc$, can be defined in a similar manner.

AMRSs are used to represent ordered collections of AFSs. However, due to the possibility of value sharing among the constituents of AMRSs, they are not sequences in the mathematical sense, and the notion of sub-structure has to be defined in order to relate them to AFSs.

**Definition 3.4 (Sub-structures)** *Let* $A = \langle Ind_A, \Pi_A, \Theta_A, \approx_A \rangle$*; let* $Ind_B$ *be a finite (contiguous) subset of* $Ind_A$*; let* $n+1$ *be the index of the first element of* $Ind_B$*. The **sub-structure** of* $A$ *induced by* $Ind_B$ *is an AMRS* $B = \langle Ind_B, \Pi_B, \Theta_B, \approx_B \rangle$ *such that:*

- $(i - n, \pi) \in \Pi_B$ *iff* $i \in Ind_B$ *and* $(i, \pi) \in A$
- $\Theta_B(i - n, \pi) = \Theta_A(i, \pi)$ *if* $i \in Ind_B$
- $(i_1 - n, \pi_1) \approx_B (i_2 - n, \pi_2)$ *iff* $i_1 \in Ind_B, i_2 \in Ind_B$ *and* $(i_1, \pi_1) \approx_A (i_2, \pi_2)$

A sub-structure of $A$ is obtained by selecting a subsequence of the indices of $A$ and considering the structure they induce. Trivially, this structure is an AMRS. We use $A^{j..k}$ to refer to the sub-structure of $A$ induced by $\{j,\ldots,k\}$. If $Ind_B = \{i\}$, $A^{i..i}$ can be identified with an AFS, denoted $A^i$.

The notion of concatenation has to be defined for AMRSs, too. Notice that by definition, concatenated AMRSs cannot share elements between them.

**Definition 3.5 (Concatenation)** *The* ***concatenation*** *of* $A = \langle Ind_A, \Pi_A, \Theta_A, \approx_A\rangle$ *and* $B = \langle Ind_B, \Pi_B, \Theta_B, \approx_B\rangle$ *of lengths* $n_A, n_B$*, respectively (denoted by* $A \cdot B$*), is an AMRS* $C = \langle Ind_C, \Pi_C, \Theta_C, \approx_C\rangle$ *such that*

- $Ind_C = \{1,\ldots,n_A+n_B\}$
- $\Pi_C = \Pi_A \cup \{(i+n_A,\pi) \mid (i,\pi)\in \Pi_B\}$
- $\Theta_C(i,\pi) = \begin{cases} \Theta_A(i,\pi) & \text{if } i \le n_A \\ \Theta_B(i-n_A,\pi) & \text{if } i > n_A \end{cases}$
- $\approx_C = \approx_A \cup \{((i_1+n_A,\pi_1),(i_2+n_A,\pi_2)) \mid (i_1,\pi_1) \approx_B (i_2,\pi_2)\}$

As usual, $A\cdot\lambda = \lambda\cdot A = A$.

We now extend the definition of unification to AMRSs: we want to allow the unification of two *AMRSs*, according to a specified set of indices. Therefore, one operand is a pair consisting of an AMRS and a set of indices, specifying some elements of it. The second operand is either an AMRS or an AFS, considered as an AMRS of length 1. Recall that due to reentrancies, other elements of the first AMRS can be affected by this operation. Therefore, the result of the unification is a new AMRS. We refer to AMRS unification as "unification in context" in the sequel to emphasize the effect that the operation might have on elements that are not directly involved in it.

**Definition 3.6 (Unification of AMRSs)** *Let* $A = \langle Ind_A, \Pi_A, \Theta_A, \approx_A\rangle$ *be an AMRS. Let* $B = \langle Ind_B, \Pi_B, \Theta_B, \approx_B\rangle$ *be an AMRS (if B is an AFS it is interpreted as an AMRS of length 1). Let* $J$ *be a set of indices such that* $J \subseteq Ind_A$*. Let* $f(i) = i$ *if B is an AMRS,* $f(i) = 1$ *if B is an AFS.* $(A,J)\sqcup B$ *is defined if B is an AMRS and* $J \subseteq Ind_B$*, or if B is an AFS and* $|J| = 1$*; in any case, it is the AMRS* $C' = Ty(Eq(Cl(\langle Ind_C, \Pi_C, \Theta_C, \approx_C\rangle)))$*, where*

- $Ind_C = Ind_A$
- $\Pi_C = \Pi_A \cup \{(i,\pi) \mid i \in J \text{ and } (f(i),\pi) \in \Pi_B$
- $\Theta_C(i,\pi) = \begin{cases} \Theta_A(i,\pi) & \text{if } i \notin J \\ \Theta_A(i,\pi) \sqcup \Theta_B(f(i),\pi) & \text{if } i \in J \text{ and } (i,\pi)\in\Pi_A \text{ and } (f(i),\pi)\in\Pi_B \\ \Theta_A(i,\pi) & \text{if } i \in J \text{ and } (i,\pi)\in\Pi_A \text{ and } (f(i),\pi)\notin\Pi_B \\ \Theta_B(f(i),\pi) & \text{if } i \in J \text{ and } (i,\pi)\notin\Pi_A \text{ and } (f(i),\pi)\in\Pi_B \end{cases}$
- $\approx_C = \approx_A \cup \{((i_1,\pi_1),(i_2,\pi_2)) \mid i_1,i_2 \in J \text{ and } (f(i_1),\pi_1) \approx_B (f(i_2),\pi_2)\}$

*The unification fails if there exists some pair* $(i,\pi)\in\Pi_{C'}$ *such that* $\Theta_{C'}(i,\pi) = \top$.

Many of the properties of AFSs, proven in the previous section, hold for AMRSs, too. In particular, if $A, B$ are AMRSs then so is $(A,J)\sqcup B$ if it is defined, $len((A,J)\sqcup B) = len(A)$ and $(A,J)\sqcup B \sqsupseteq A$. Also, for every two AMRSs $A, B$, $(A,\{1\ldots len(A)\})\sqcup B = A$ iff $B^{1..len(A)} \sqsubseteq A$.

# 4 Parsing

Parsing is the process of determining whether a given string belongs to the language defined by a given grammar, and assigning a structure to the permissible strings. Various parsing algorithms exist for various classes of grammars. In this section we formalize and explicate some of the notions of [3, chapter 13]. We give direct definitions for rules, grammars and languages, based on our new notion of AMRSs. This presentation is more adequate to current TFS-based systems than [7, 17], that use first-order terms. Moreover, it does not necessitate special, ad-hoc features and types for encoding trees in TFSs as [16] does. We don't assume any explicit context-free back-bone for the grammars, as do [10] or [18].

The parsing algorithm we describe is a pure bottom-up one that makes use of a chart to record edges. The formalism we presented is aimed at being a platform for specifying grammars in HPSG, which is characterized by employing a few very general rules (or rule schemata); selecting the rules that are applicable in every step of the process requires unification anyhow. Therefore we choose a particular parsing algorithm that does not make use of top down predictions but rather assumes that every rule might be applied in every step. This assumption is realized by initializing the chart with predictive edges for every rule, in every position.

## 4.1 Rules and Grammars

We define rules and grammars over a fixed set WORDS of words (in addition to the fixed sets FEATS and TYPES). We use $w$ to refer to elements of WORDS, $w_i$ to refer to strings over WORDS. We assume that the lexicon associates with every word $w_i$ a set of feature structures $Cat(w_i)$, its **category**,[1] so we can ignore the terminal words and consider only their categories. The input for the parser, therefore, is a sequence[2] of sets of TFSs rather than a string of words.

**Definition 4.1 (Pre-terminals)** *Let $w = w_1 \ldots w_n \in$ WORDS$^*$. $PT_w(j,k)$ is defined iff $1 \leq j,k \leq n$, in which case it is the set of AMRSs $Abs(\langle A_j, A_{j+1}, \ldots, A_k \rangle)$ where $A_i \in Cat(w_i)$ for $j \leq i \leq k$. If $j > k$ then $PT_w(j,k) = \{\lambda\}$. We omit the subscript $w$ when it is clear from the context.*

**Lemma 4.2** *If $w = w_1 \cdots w_n$, $1 \leq i \leq j \leq k \leq n$, $A \in PT_w(i,j)$ and $B \in PT_w(j+1,k)$ then $A \cdot B \in PT_w(i,k)$.*

**Proof:** An immediate corollary of the definition.

**Definition 4.3 (Rules)** *A **rule** is a MRS of length $n > 0$ with a distinguished last element. If $\langle A_1, \ldots, A_{n-1}, A_n \rangle$ is a rule then $A_n$ is its **head**[3] and $\langle A_1, \ldots, A_{n-1} \rangle$ is its **body**.[4] We write such a rule as $\langle A_1, \ldots, A_{n-1} \Rightarrow A_n \rangle$. In addition, every category of a lexical item is a rule (with an empty body). We assume that such categories don't head any other rule.*

Notice that the definition supports $\epsilon$-rules, i.e., rules with null bodies.

**Definition 4.4 (Grammars)** *A **grammar** $G = (\mathcal{R}, A_s)$ is a finite set of rules $\mathcal{R}$ and a **start symbol** $A_s$ that is a TFS.*

An example grammar, whose purpose is purely illustrative, is depicted in figures 6 and 7 in the appendix. A discussion of the methodological status of the start symbol appears later on in this section, prior to the definition of *languages*.

For the following discussion we fix a particular grammar $G = (\mathcal{R}, A_s)$. We define a *derivation* relation over AMRSs as the basis for defining the *language* of TFS-based grammars. Checking whether two given AMRSs $A$ and $B$ stand in the derivation relation is accomplished by the following steps: first, an element of $A$ has to be selected; this element has to unify with the head of some rule $\rho$; then, a sub-structure of $B$ is selected; this substructure has to unify with the body of $\rho$. All unifications are done in context, so that other components of the AMRSs involved may be affected, too. Moreover, there must be some way to record the effects of successive unifications; to this end, derivation is defined only for pairs of AMRSs that are already "as specific as needed"; that is to say, if the rule adds any information to the AMRSs, this information already has to be recorded in them in order for them to be related by derivation. This is why, in the definition below, we use an AMRS $R$ that is *at least as specific as* some rule $\rho$, and not $\rho$ itself, to guide the derivation. This is also why the definition requires that all the unifications do not add information. *strong derivation* is the relation that holds between such AMRSs; another relation, *derivation*, relaxes that requirement by allowing two AMRSs to be related even if they contain only part of the information that is required for strong derivation to hold.

[1] $Cat(w_i)$ is a singleton if $w_i$ is unambiguous.
[2] We assume that there is no reentrancy among lexical items.
[3] This use of *head* must not be confused with the linguistic one, the core features of a phrase.
[4] Notice that the traditional direction is reversed and that the head and the body need not be disjoint.

Since elements of AMRSs involve indices that denote their linear position in the sequence of roots, the operation of *replacing* some element in one AMRS with a sub-structure, whose length might be greater than one, becomes notationally complicated. Conceptually, though, it resembles very much the replacement of some symbol with a sequence of symbols in context-free derivation, or the replacement of the selected goal (that unifies with the head of some rule) with the body of the rule, in Prolog SLD-resolution. One main difference in our definition is that we do not carry substitutions through sequences of derivations; rather, we treat all the pairs in a derivation sequence as if the appropriate substitutions have already been applied to them (recall that members of these pairs are "as specific as needed").

**Definition 4.5 (Strong Derivation)** *An AMRS $A = \langle Ind_A, \Pi_A, \Theta_A, \approx_A\rangle$ of length $k$* ***strongly derives*** *an AMRS $B$ (denoted $A \rightarrow B$) iff*

- *there exist a rule $\rho \in \mathcal{R}$ and an AMRS $R \sqsupseteq Abs(\rho)$ (with $len(R) = n$), such that:*
- *some element of $A$ unifies with the head of $R$, and some sub-structure of $B$ unifies with the body of $R$; namely, there exist $j \in Ind_A$ and $i \in Ind_B$ such that:*
  $A = (A, \{j\}) \sqcup R^n$, $\quad B^{i\ldots i+n-2} = (B, \{i \ldots i+n-2\}) \sqcup R^{1\ldots n-1}$,
  $R = (R, \{n\}) \sqcup A^j$, $\quad R = (R, \{1 \ldots n-1\}) \sqcup B^{i\ldots i+n-2}$
- *$B$ is the replacement of the $j$-th element of $A$ with the body of $R$; namely, let*

$$f(i) = \begin{cases} i & \text{if } 1 \leq i < j \\ i+n-2 & \text{if } j < i \leq k \end{cases}, \qquad g(i) = i+j-1 \text{ if } 1 \leq i < n$$

  *then $B = Ty(Eq(Cl(\langle Ind_{B'}, \Pi_{B'}, \Theta_{B'}, \approx_{B'}\rangle)))$, where*

  - $Ind_{B'} = \langle 1, \ldots, k+n-2\rangle$
  - $(i, \pi) \in \Pi_{B'}$ *iff* $\begin{cases} i = f(i') \text{ and } (i', \pi) \in \Pi_A \text{ or} \\ i = g(i') \text{ and } (i', \pi) \in \Pi_R \end{cases}$
  - $\Theta_{B'}(i, \pi) = \begin{cases} \Theta_A(i', \pi) & \text{if } i = f(i') \\ \Theta_R(i', \pi) & \text{if } i = g(i') \end{cases}$
  - $(i_1, \pi_1) \approx_{B'} (i_2, \pi_2)$ *if*
    * $i_1 = f(i'_1)$ *and* $i_2 = f(i'_2)$ *and* $(i'_1, \pi_1) \approx_A (i'_2, \pi_2)$, *or*
    * $i_1 = g(i'_1)$ *and* $i_2 = g(i'_2)$ *and* $(i'_1, \pi_1) \approx_R (i'_2, \pi_2)$, *or*
    * $i_1 = f(i'_1)$ *and* $i_2 = g(i'_2)$ *and there exist* $\pi_1, \pi_2, \pi_3$ *such that* $(i'_1, \pi_1) \approx_A (j, \pi_3)$ *and* $(n, \pi_3) \approx_R (i'_2, \pi_2)$, *or*
    * $i_1 = g(i'_1)$ *and* $i_2 = f(i'_2)$ *and there exist* $\pi_1, \pi_2, \pi_3$ *such that* $(i'_1, \pi_1) \approx_R (j, \pi_3)$ *and* $(n, \pi_3) \approx_A (i'_2, \pi_2)$

*The reflexive transitive closure of '$\rightarrow$', denoted '$\xrightarrow{*}$', is defined as follows: $A \xrightarrow{*} A''$ if $A = A''$ or if there exists $A'$ such that $A \rightarrow A'$ and $A' \xrightarrow{*} A''$.*

Intuitively, $A$ strongly derives $B$ through some AFS $A^j$ in $A$, if some rule $\rho \in \mathcal{R}$ licenses the derivation. $A^j$ is unified with the head of the rule, and if the unification succeeds, the (possibly modified) body of the rule replaces $A^j$ in $B$. The definition is graphically demonstrated in figure 4.

**Lemma 4.6** *If $A \rightarrow B$ and $A \sqsubseteq A'$ then there exists $B'$ such that $B \sqsubseteq B'$ and $A' \rightarrow B'$.*

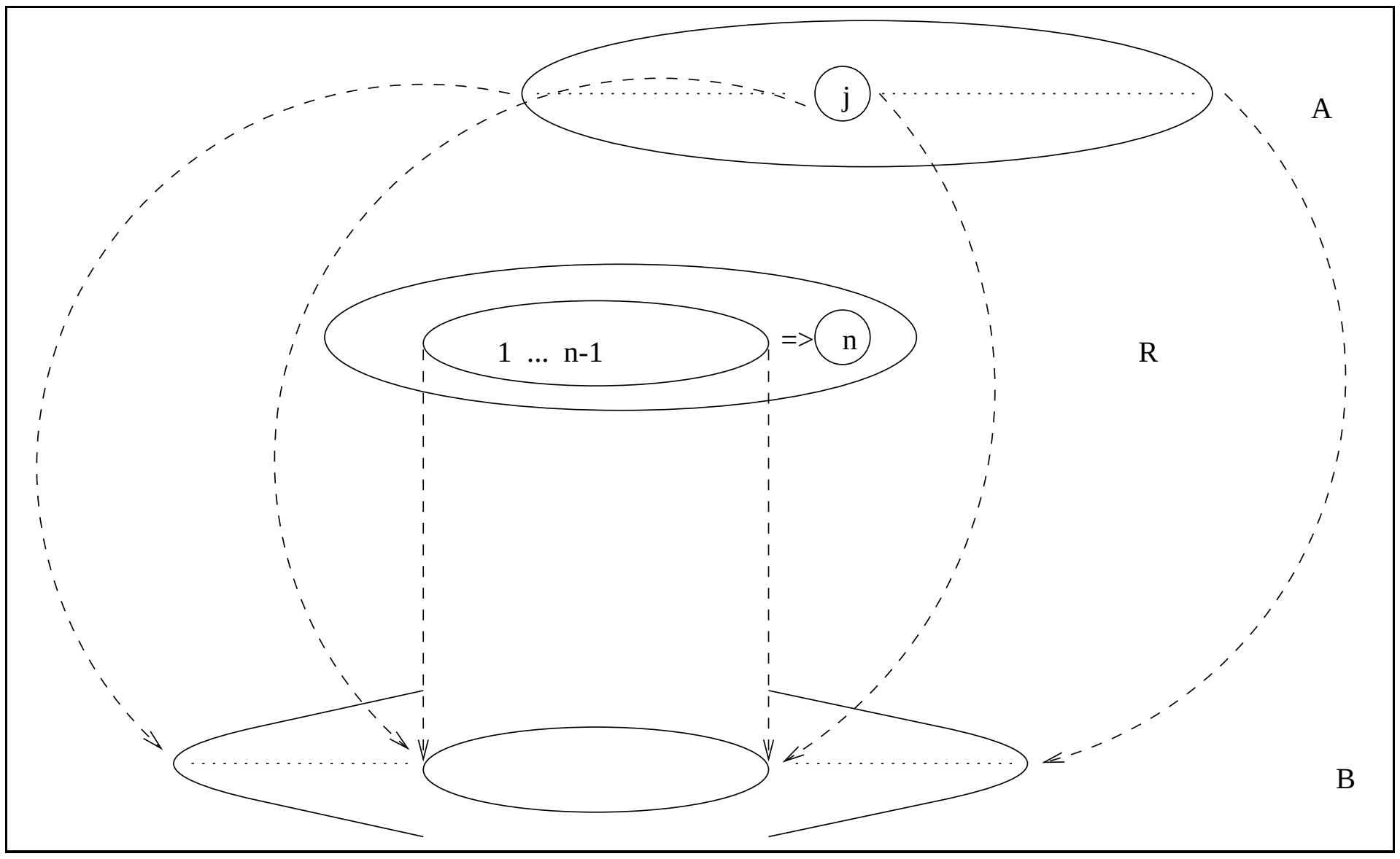

Figure 4: Strong derivation

**Proof:** $A \to B$, therefore there exists a rule $\rho \in \mathcal{R}$, an AMRS $R \sqsupseteq Abs(\rho)$ and an index $j$ such that $A$ unifies with the head of $R$, and $B$ is obtained by replacing $A^j$ with the body of $R$. $A$ and $B$ are already "as specific as needed"; thus, since $A \sqsubseteq A'$ and $A = (A, \{j\}) \sqcup R^n$, $A' = (A', \{j\}) \sqcup R^n$. Hence there exists $R' \sqsupseteq R$ such that $R' = (R', \{n\}) \sqcup A'^j$, $A'$ unifies with its head and $B'$ is obtained by replacing the $j$-th element of $A'$ with its body.

**Lemma 4.7** *If $A \xrightarrow{*} B$ and $A \sqsubseteq A'$ then there exists $B'$ such that $B \sqsubseteq B'$ and $A' \xrightarrow{*} B'$.*

**Proof:** By induction on the derivation sequence and lemma 4.6.

**Lemma 4.8** *If $A^{1...k} \xrightarrow{*} B$ and $A^{k+1} \xrightarrow{*} C$ then $A^{1...k+1} \xrightarrow{*} B \cdot C$.*

**Proof:** The derivation is obtained by applying first the derivation steps that derive $B$ from $A^{1...k}$ and then those that derive $C$ from $A^{k+1}$. Since $A^{1...k} \xrightarrow{*} B$, $A$ is "as specific as needed" and the application of the derivation steps from $A$ to $B$ does not affect the applicability of the derivations step to $C$.

**Lemma 4.9** *If $A \sqsupseteq Abs(\rho)$ for some $\rho \in \mathcal{R}$ of length $n$ then $A^n \to A^{1...n-1}$.*

**Proof:** Immediate from the definition of derivation.

There are various definitions in the literature for the *language* that is defined by a grammar $G$ expressed in a unification-based grammar formalism. For example, [15, 16] do not include a start symbol in the grammar at all, and define $L(G)$ as the set of strings derivable from *some* feature structure. In [17] a start symbol is defined (notated *goal axiom*), and $L(G)$ is defined as the set of strings that are derivable from some generalization of the start symbol, i.e., from some feature structure that subsumes it. [18], on the other hand, assumes that a specific feature *cat* is present in every feature structure (the value of which simulates non-terminals in a context-free "underlying" grammar), and uses this feature to single out the start symbol: $L(G)$ is the set of strings that are derivable from some feature structure in which the *cat* feature is $S$ (the start symbol of the underlying context-free grammar). A similar definition is given by [7]: $L(G)$ is the set of strings derivable from the start symbol, where the start symbol is a *constant* (that is, an atomic feature structure).

There is a good motivation to employ a start symbol: the grammar writer might want to specify a certain criterion for the permissible strings in the language, for example, that they are all *sentences*. Moreover, it makes sense to include in the language such strings that are not derived directly by the start symbol, but rather by a TFS that is related to the start symbol. For example, the grammar writer might state that only TFSs with a *cat* feature valued $S$ are permissible, meaning that every TFS that *is subsumed* by the start symbol (that is, contains all the information it encodes) is a sentence. However, such a definition prevents the incorporation of *subsumption test* (see section 4.3.3 below) into the parsing, since the correctness of the computation can not be maintained.

Due to these consideration we chose a relaxed condition on the start symbol in our definition of languages. We define a derivation relation between AMRSs in a way that allows the initial symbol of the grammar to derive a sequence of lexical entries even if the actual (strong) derivation is between a TFS that unifies with the start symbol and a more specific instance of the pre-terminals.

**Definition 4.10 (Derivation)** *An AMRS $A$ **derives** an AMRS $B$ ($A \leadsto B$) iff there exist AMRSs $A', B'$ such that $(A, \{1, \ldots, len(A)\}) \sqcup A' \neq \top$, $B \sqsubseteq B'$ and $A' \xrightarrow{*} B'$.*

**Definition 4.11 (Language)** *The **language** of a grammar $G$ is $L(G) = \{w = w_1 \cdots w_n \in$ WORDS$^* \mid Abs(A_s) \leadsto B$ for some $B \in PT_w(1, n)\}$.*

Figure 5 shows a sequence of derivation steps, starting from some feature structure that is more specific than the initial symbol and ending in a sequence of structures that can stand for the string "John loves fish", based upon the example grammar. While we use identical tags within different MRSs it must be understood that the scope of tags is limited to a single AMRS. When two AMRSs are related by derivation, there can be no reentrancies between them. The use of identical tags here is for explanatory reasons only.

## 4.2 Parsing as Operational Semantics

We view parsing as a computational process, capable of endowing TFS formalisms with an operational semantics, which can be used to derive control mechanisms for an abstract machine we design (see [19]).

As is well known (see, e.g., [11]), the meaning of a logic program $P$ can be specified algebraically as the least fix-point (lfp) of the *immediate consequence* operator $T_P$ of the program. A similar approach can be applied to a context-free grammar $G$, such that $L(G)$ equals (a projection of) the least fix-point of an analogous *immediate derivation* operator, $T_G$. Let $G = (V, T, P, S)$ be a context-free grammar.[5] Let $I \subseteq V \times T^*$. Define $T_G(I) = \{\langle A, w\rangle \mid A \rightarrow w \in P, w \in T\} \cup \{\langle A, w_1 \cdots w_k\rangle \mid A \rightarrow A_1 \cdots A_k \in P, \langle A_i, w_i\rangle \in I, 1 \leq i \leq k\}$. Then the least fix-point of $T_G$ is the union over $A \in V$ of $\{\langle A, w\rangle \mid w \in L_A(G)\}$.

In a sense, computing the lfp of $T_G$ corresponds to computing the language generated by $G$. Parsing, then, amounts to checking if the input $w$ is in the language. This process induces an inherently inefficient computation: since $w$ is given, it can be used to optimize the computation. This is achieved by defining $T_{G,w}$, a *parsing step* operator, which is dependent on the input sentence $w$. The set of items $I$ has to be extended, too: an item is a triple $[i, A, j]$ where $0 \leq i, j, \leq n$ ($n$ being the length of $w$) and $A \in V$. Informally, an item $[i, A, j]$ represents the existence of a derivation for the symbol $A$ to a substring of $w$, namely $w_i \ldots w_j$. $w \in L_S(G)$ if and only if $[1, S, n] \in lfp(T_{G,w})$, so that parsing now amounts to computing the least fixed point of $T_{G,w}$, which is more efficient, and then checking whether the appropriate item is in the lfp.

We now return to TFS-based formalisms and define $T_{G,w}$ for a TFS-based grammar $G$, thus providing means for defining the meaning of $G$. A computation is triggered by some input string of words $w = w_1 \cdots w_n$ of length $n \geq 0$. For the following discussion we fix a particular grammar $G = (\mathcal{R}, A_s)$ and a particular input string $w$ of length $n$. A *state* of the computation is a set of *items*, and states are related by a transition relation. The presentation below corresponds to a pure bottom-up parsing algorithm, as it is both simple and efficient.

[5] We assume a normal form, where for $A \rightarrow \alpha \in P$, either $\alpha \in T$ or $\alpha \in V^*$.

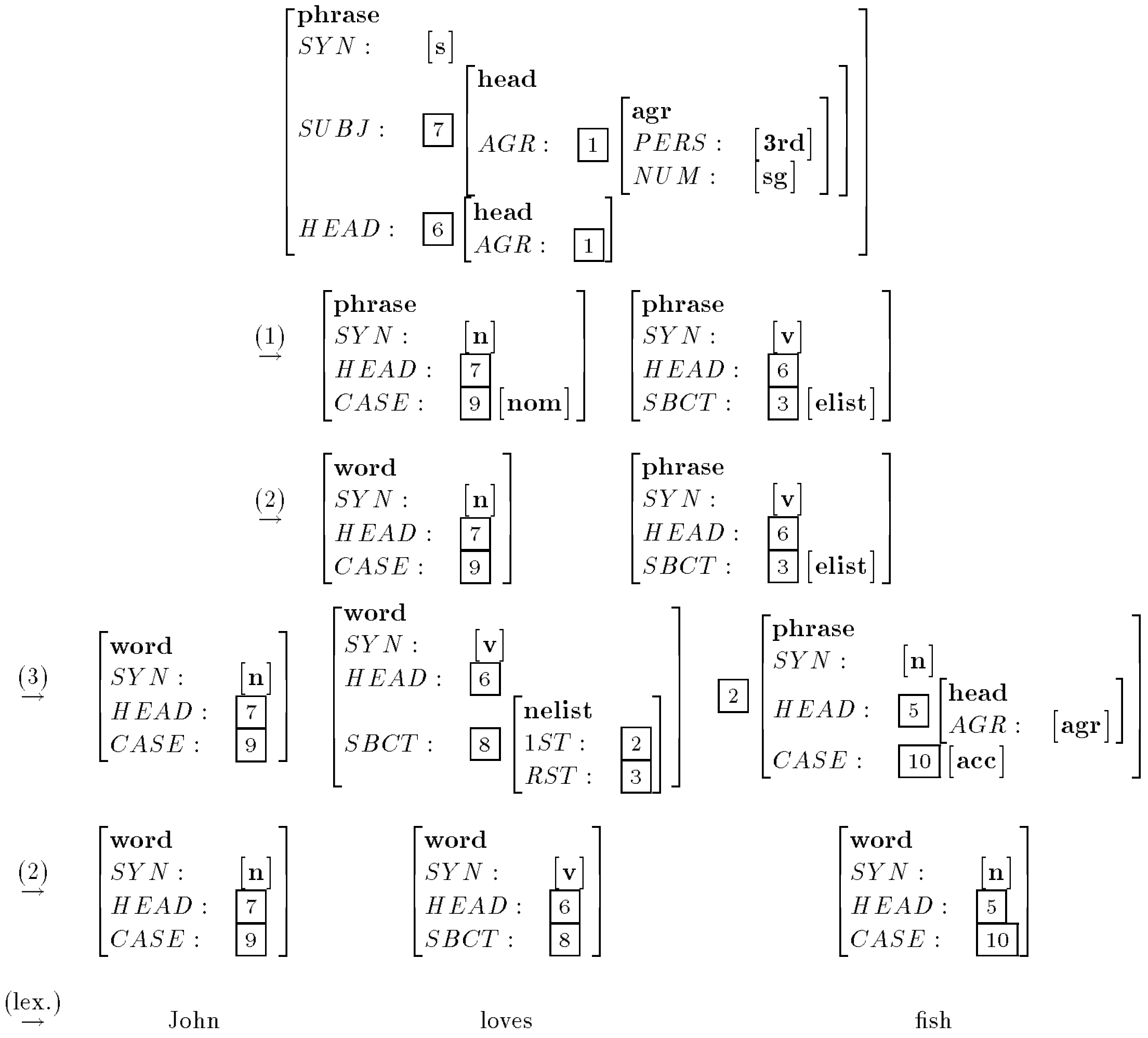

Figure 5: A leftmost derivation

**Definition 4.12 (Items)** *An* ***item*** *is a four-tuple* $[i, A, j, c]$*, where* $i, j \in I\!N$*,* $i \leq j$*,* $A$ *is an AMRS and* $c$ *is either* Act*, in which case the item is* ***active****, or* Comp*, in which case it is* ***complete****. Let* Items *be the collection of all items.*

If $[i, A, j, c]$ is an item, we say that $A$ *spans* the input from position $i+1$ to position $j$ (inclusive). $A$ can be seen as a representation of a *dotted rule*, or *edge*: during parsing all generated items are such that $A$ is (possibly more specific than) a prefix of some grammar rule. The notion of items usually employs edges that contain entire rules, whereas we only use prefixes of rules. This difference is not essential, and in an actual implementation of a parser that is induced by $T_{G,w}$, edges indeed include a reference to the rule on which they rely.

In what follows we defined $T_{G,w}$, a parsing operator that corresponds (bottom-up) chart parsing. However, it is possible to characterize algebraic operators that correspond to other parsing schemas as well.

**Definition 4.13** *Let* $T_{G,w} : 2^{\text{Items}} \rightarrow 2^{\text{Items}}$ *be a transformation on sets of items, where for*

$I \in$ ITEMS, $x \in T_{G,w}(I)$ *iff either*

$$\begin{array}{l} \exists \rho \in \mathcal{R}, Abs(\rho) = R = A_1, \ldots, A_{m-1} \Rightarrow A_m, m > 1 \\ \exists k < m-1 \\ \exists \alpha \in I, \alpha = [i_\alpha, A_\alpha, j_\alpha, \textsc{Act}], len(A_\alpha) = k \\ \exists \beta \in I, \beta = [i_\beta, A_\beta, j_\beta, \textsc{Comp}], len(A_\beta) = 1 \\ j_\alpha = i_\beta \\ B = (R, \{1 \ldots k\}) \sqcup A_\alpha \\ C = (B, \{k+1\}) \sqcup A_\beta \\ x = [i_\alpha, C^{1 \ldots k+1}, j_\beta, \textsc{Act}] \end{array} \tag{1}$$

*or*

$$\begin{array}{l} \exists \rho \in \mathcal{R}, Abs(\rho) = R = A_1, \ldots, A_{m-1} \Rightarrow A_m, m > 1 \\ \exists \alpha \in I, \alpha = [i_\alpha, A_\alpha, j_\alpha, \textsc{Act}], len(A_\alpha) = m-1 \\ C = (R, \{1 \ldots m-1\}) \sqcup A_\alpha \\ x = [i_\alpha, C^m, j_\alpha, \textsc{Comp}] \end{array} \tag{2}$$

*or*

$$\begin{array}{l} \exists i, 0 \le i \le n \\ x = [i, \lambda, i, \textsc{Act}] \end{array} \tag{3}$$

*or*

$$\begin{array}{l} \exists \rho \in \mathcal{R}, len(\rho) = 1 \\ \exists i, 0 \le i \le n \\ x = [i, Abs(\rho), i, \textsc{Comp}] \end{array} \tag{4}$$

*or*

$$\begin{array}{l} w = w_1, \ldots, w_n, n \ge 1 \\ \exists i, 0 < i \le n \\ x = [i-1, Abs(A_i), i, \textsc{Comp}], A_i \in Cat(w_i) \end{array} \tag{5}$$

Cases 1 and 2 perform the operation known as *completion*: 1 moves the dot one position along the body of a rule, and 2 creates a complete item once the dot reaches the end of the body. Case 3 corresponds to the *prediction* operation, whereas case 5 corresponds to *scanning*. Case 4 handles $\epsilon$-rules, i.e., rules with null bodies, and creates complete items that span a null substring of the input sentence. Notice that cases 3 and 4 are independent of the argument $I$ and therefore add the same items in every application of $T_{G,w}$. Case 5 is also independent of the argument, but is dependent on the input sentence $w$.

The operator $T_{G,w}$, on which the algebraic semantics of TFS-based grammars is based, naturally induces an operational semantics for such formalisms: once the operator is shown to be continuous, a computational process that corresponds to the iterative application of $T_{G,w}$ computes the set of items in the least fix-point of the operator. This process can be thought of as an analog of bottom-up, chart-based parsing: the chart is initialized with predictions for every rule in every position (by operation 3) and with complete edges for every input word (by operation 5). Then, operations 1 and 2 are used to apply the grammar rules using the chart, in an unspecified order. We prove below that the process is indeed analogous to parsing $w$ with respect to $G$.

**Theorem 4.14** *$T_{G,w}$ is monotone: if $I_1 \subseteq I_2$ then $T_{G,w}(I_1) \subseteq T_{G,w}(I_1)$.*

**Proof:** Suppose $I_1 \subseteq I_2$. If $x \in T_{G,w}(I_1)$ then $x$ was added by one of the five operations; operations 3, 4 and 5 add the same items every time $T_{G,w}$ is applied, and thus $x \in T_{G,w}(I_2)$, too. If $x$ was added by operation 1, then there exist items $\alpha, \beta$ in $I_1$ to which this operation applies. Since $I_1 \subseteq I_2$, $\alpha, \beta$ are in $I_2$, too, and hence $x \in T_{G,w}(I_2)$, too. The same applies for operation 2. In any case, $x \in T_{G,w}(I_2)$ and hence $T_{G,w}(I_1) \subseteq T_{G,w}(I_2)$.

**Theorem 4.15** *$T_{G,w}$ is continuous: if $I_i, i \ge 0$ is a chain, then $T_{G,w}(\bigcup_i I_i) = \bigcup_i T_{G,w}(I_i)$.*

**Proof:** First, $T_{G,w}$ is monotone. Second, let $I = I_0 \subseteq I_1 \subseteq \ldots$ be a chain of items. If $x \in T_{G,w}(\bigcup_{i \ge 0} I_i)$ then there exist $\alpha, \beta \in \bigcup_{i \ge 0} I_i$ as required, due to which $x$ is added. Then there exist $i, j$ such that $\alpha \in I_i$ and $\beta \in I_j$. Let $k$ be the maximum of $i, j$. Then $\alpha, \beta \in I_k$, $x \in T_{G,w}(I_k)$ and hence $x \in \bigcup_{i \geq 0} T_{G,w}(I_i)$.
If $x \in \bigcup_{i \geq 0} T_{G,w}(I_i)$ then there exists some $i$ that $x \in T_{G,w}(I_i)$. $I_i \subseteq \bigcup_{i \geq 0} I_i$ and since $T_{G,w}$ is monotone, $T_{G,w}(I_i) \subseteq T_{G,w}(\bigcup_{i \geq 0} I_i)$, and hence $x \in T_{G,w}(\bigcup_{i \geq 0} I_i)$. Therefore $T_{G,w}$ is continuous.

**Corollary 4.16** *The least fix-point of $T_{G,w}$ can be obtained by iteratively computing $I_{m+1} = T_{G,w}(I_m)$, starting from $I_0 = \phi$ and stopping when a fix-point is reached.*

**Proof:** By Tarski-Knaster theorem, the lfp exists since $T_{G,w}$ is monotone; By Kleene's theorem, since $T_{G,w}$ is continuous, the lfp can be obtained by applying the operator iteratively, starting from $\phi$.

**Definition 4.17 (Algebraic meaning)** *The **meaning** of a grammar $G$ with respect to an input sentence $w$ is the least fix-point of the operator $T_{G,w}$.*

**Definition 4.18 (Computation)** *The **$w$-computation** triggered by $w \in \text{WORDS}^*$ is the infinite sequence of sets of items $I_i, i \geq 0$, such that $I_0 = \phi$ and for every $m \geq 0$, $I_{m+1} = T_{G,w}(I_m)$. The computation is **terminating** if there exists some $m \geq 0$ for which $I_m = I_{m+1}$ (i.e., a fix-point is reached in finite time). The computation is **successful** if there exists some $m$ such that $[0, A, n, \text{COMP}] \in I_m$, where $len(A) = 1$ and $A \sqcup Abs(A_s) \neq \top$; otherwise, the computation fails.*

Notice that we check whether the generated items are *unifiable* with the initial symbol, in accordance with the definition of *languages*. If the initial symbol of the grammar is interpreted differently when languages are defined, a corresponding modification has to be made in the condition for *success*.

An example of parsing with respect to the example grammar given in figure 7 is given in the appendix.

## 4.3 Proof of Correctness

In this section we show that parsing, as defined above, is (partially) correct. First, the algorithm is *sound*: a $w$-computation succeeds only if $w \in L(G)$; second, it is *complete*: if $w \in L(G)$, it triggers a successful $w$-computation. Computations are not guaranteed to terminate, but we show that termination is assured for a certain subset of the grammars that are *off-line parsable*. We discuss off-line parsability in section 4.3.4.

### 4.3.1 Soundness

In what follows we fix a particular $w$-computation $I_0, I_1, \ldots$, triggered by some input $w = w_1 \cdots w_n$.

**Lemma 4.19** *If $[i, A, j, \text{COMP}] \in I_l$ for some $l$ then $len(A) = 1$.*

**Proof:** By definition of $T_{G,w}$, complete items are generated by operations 2, 3 and 4. All these operations add items in which the AMRS is of length 1.

**Lemma 4.20** *If $[i, A, j, \text{ACT}] \in I_l$ for some $l$ and $len(A) = k > 0$ then there exists $\rho \in \mathcal{R}$ such that $Abs(\rho)^{1 \ldots k} \sqsubseteq A$.*

**Proof:** By induction on $l$. If $l = 0$ then $I_l = \phi$ and the proposition holds vacuously. Assume that the proposition holds for every $l' < l$. Suppose that $x = [i, A, j, \text{ACT}] \in I_l$ and $A \neq \lambda$. Then $x$ must have been added by operation 1 (dot movement). Then $x = [i_\alpha, C^{1 \ldots k+1}, j_\beta, \text{ACT}]$ where $C = ((R, \{1 \ldots k\}) \sqcup A_\alpha), \{k+1\}) \sqcup A_\beta$, namely $C \sqsupseteq R$ and thus $C^{1 \ldots k+1} \sqsupseteq R^{1 \ldots k+1}$.

**Theorem 4.21 (Parsing invariant (a))** *If $[i, A, j, c] \in I_l$ and $i < j$ then there exist $B \in PT_w(i+1, j)$ and $A' \sqsupseteq B$ such that $A \xrightarrow{*} A'$. If $i = j$ then $A \xrightarrow{*} \lambda$.*

**Proof:** By induction on $l$.
If $l = 0$ then $I_l = \phi$ and the proposition holds vacuously.
Assume that the proposition holds for every $l' < l$. Suppose that $x = [i, A, j, c] \in I_l$. Then $x$ must have been added by one of the operations. Consider each case separately:

1. **dot movement:** $x = [i_\alpha, C^{1\ldots k+1}, j_\beta, \textsc{Act}]$ where there exist $\alpha, \beta \in I_{l-1}$ as required and $C = (B, \{k+1\}) \sqcup A_\beta$, $B = (R, \{1 \ldots k\}) \sqcup A_\alpha$. By the induction hypothesis, there exist $A'_\alpha, B_\alpha$ such that $A_\alpha \xrightarrow{*} A'_\alpha$ and $A'_\alpha \sqsupseteq B_\alpha \in PT(i_\alpha + 1, j_\alpha)$. Also, there exist $A'_\beta, B_\beta$ such that $A_\beta \xrightarrow{*} A'_\beta$ and $A'_\beta \sqsupseteq B_\beta \in PT(j_\alpha + 1, j_\beta)$. $B^{1\ldots k} = (R, \{1 \ldots k\}) \sqcup A_\alpha$; if $k > 0$, $A_\alpha \neq \lambda$ and by lemma 4.20 $A_\alpha \sqsupseteq R$, hence $B^{1\ldots k} = A_\alpha$. If $k = 0$, $B^{1\ldots k} = \lambda = A_\alpha$. Hence $B^{1..k} \xrightarrow{*} A'_\alpha$. $C^{1..k} \sqsupseteq B^{1..k}$, and by lemma 4.7 there exists $A''_\alpha \sqsupseteq A'_\alpha$ such that $C^{1..k} \xrightarrow{*} A''_\alpha$. In the same way, there exists $A''_\beta \sqsupseteq A'_\beta$ such that $C^{k+1} \xrightarrow{*} A''_\beta$. By lemma 4.8, $C^{1\ldots k+1} \xrightarrow{*} A''_\alpha \cdot A''_\beta$. But $A''_\alpha \cdot A''_\beta \sqsupseteq A'_\alpha \cdot A'_\beta \sqsupseteq B_\alpha \cdot B_\beta$, and since $B_\alpha \in PT(i_\alpha + 1, j_\alpha)$ and $B_\beta \in PT(j_\alpha + 1, j_\beta)$, by lemma 4.2 $B_\alpha \cdot B_\beta \in PT(i_\alpha + 1, j_\beta)$. The cases in which $i_\alpha = j_\alpha$ or $i_\beta = j_\beta$ are trivial.

2. **completion:** $x = [i_\alpha, C^m, j_\alpha, \textsc{Comp}]$ where $C = (R, \{1 \ldots m-1\}) \sqcup A_\alpha$ and there exist an abstract rule $R$ and an item $\alpha \in I_{l-1}$ as required, and (by lemma 4.20) $A_\alpha^{1\ldots m-1} \sqsupseteq R^{1\ldots m-1}$. If $i_\alpha < j_\alpha$ then by the induction hypothesis, there exist $A'_\alpha, B_\alpha$ such that $A_\alpha \xrightarrow{*} A'_\alpha$ and $A'_\alpha \sqsupseteq B_\alpha \in PT(i_\alpha + 1, j_\alpha)$. $C = (R, \{1 \ldots m-1\}) \sqcup A_\alpha$, hence $C^{1\ldots m-1} = A_\alpha$ and thus $C^{1\ldots m-1} \xrightarrow{*} A_\alpha$. From lemma 4.9, $C^m \rightarrow C^{1\ldots m-1}$, and thus $C^m \xrightarrow{*} A'_\alpha$. If $i_\alpha = j_\alpha$ then $A_\alpha \xrightarrow{*} \lambda$ and hence $C^m \xrightarrow{*} \lambda$.

3. **prediction:** $x = [i, \lambda, i, \textsc{Act}]$ and $PT(i+1, i) = \phi$.

4. **$\epsilon$-rules:** $x = [i, Abs(\rho), i, \textsc{Comp}]$ and $PT(i+1, i) = \phi$.

5. **scanning:** $x = [i-1, Abs(A_i), i, \textsc{Comp}]$ where $A_i \in Cat(w_i)$, and $Abs(A_i) \xrightarrow{*} Abs(A_i)$ trivially. $Abs(A_i) \in PT(i+1, j)$ by definition.

**Theorem 4.22** *If a computation, triggered by $w$, is successful, then $w \in L(G)$.*

**Proof:** If a computation is successful then there exists some $m \geq 0$ such that $x = [0, A, n, \textsc{Comp}] \in I_m$ where $len(A) = 1$ and $A \sqcup Abs(A_s) \neq \top$. By the parsing invariant, $A \xrightarrow{*} A'$ for some $A' \sqsupseteq B \in PT_w(1, n)$. Hence $Abs(A_s) \leadsto B$ and $w \in L(G)$.

### 4.3.2 Completeness

The following theorem shows that one derivation step, licensed by a rule $Abs(\rho)$ of length $r$, corresponds to $r+1$ applications of $T_{G,w}$, starting with an item that predicts the rule and advancing the dot $r$ times, until a complete item for that rule is generated.

**Theorem 4.23 (Parsing invariant (b))** *If $A \xrightarrow{*} A'$ and $A' \sqsupseteq B \in PT_w(i+1, j)$ then for every $k$, $0 < k \leq len(A)$, there exists $l_k$ such that $[i_k, C_k, j_k, \textsc{Comp}] \in I_{l_k}$, where $C_k \sqsubseteq A^k$, $i_1 = i$, $j_{len(A)} = j$ and $j_k = i_{k+1}$ if $k < len(A)$.*

**Proof:** By induction on $l$, the number of derivation steps from $A$ to $A'$:
If $l = 0$, $A = A' \sqsupseteq B$. Since $B \in PT_w(i+1, j)$, $B = Abs(A_{i+1}) \cdot \ldots \cdot Abs(A_j)$ where $A_k \in Cat(w_k)$ for $i+1 \leq k \leq j$. The scanning operation (5) of $T_{G,w}$ adds appropriate items whenever it is applied.
Assume that $A \rightarrow D \xrightarrow{*} B \sqsupseteq PT_w(i+1, j)$ and the proposition holds for $D$ and $B$. By the induction hypothesis, for every $k$, $0 < k \leq len(D)$, there exists $l_k$ such that $[i_k, C_k, j_k, \textsc{Comp}] \in I_{l_k}$, where $C_k \sqsubseteq A^k$. Suppose that $A \rightarrow D$ through a rule $\rho$ of length $r$ by expanding $A^y$ to $D^{x\ldots x+r-1}$. Then the following sequence of items is generated, where for every $m$, $C_{1\ldots m} \sqsubseteq D^{x\ldots x+m-1}$, and $C_r \sqsubseteq A^y$:

| | | | |
|---|---|---|---|
| $[i, \lambda, i, \textsc{Act}]$ | $\in$ | $I_1$ | by prediction (3) |
| $[i_1, C_1, j_1, \textsc{Comp}]$ | $\in$ | $I_{l_1}$ | by the induction hypothesis |
| $[i, C_1, j_1, \textsc{Act}]$ | $\in$ | $I_{l_1}$ | by dot movement (1) |
| $[i_2, C_2, j_2, \textsc{Comp}]$ | $\in$ | $I_{l_2}$ | by the induction hypothesis |
| $[i, C_{1\ldots 2}, j_2, \textsc{Act}]$ | $\in$ | $I_{max(l_1,l_2)}$ | by dot movement (1) |
| $\vdots$ | | | |
| $[i, C_{1\ldots r-1}, j_{r-1}, \textsc{Act}]$ | $\in$ | $I_{max(l_1,\ldots,l_{r-1})}$ | by dot movement (1) |
| $[i, C_r, j, \textsc{Comp}]$ | $\in$ | $I_{max(l_1,\ldots,l_{r-1})+1}$ | by completion (2) |

Items are generated by the dot movement (1) operation since the conditions for its application obtain: it is easy to see that the indices $(i, j)$ match; in addition, if for some $m$, $C_{1\ldots m} \sqsubseteq D^{x\ldots x+m-1}$, and for every $k$, $C_k \sqsubseteq A^k$, then there exists $C_{1\ldots m+1}$ that is obtained by unifying some $R \sqsupseteq Abs(\rho)$ first with $C_{1\ldots m}$ and then with $C^{m+1}$, such that $C_{1\ldots m+1} \sqsubseteq D^{x\ldots x+m}$ as required. Therefore, by induction on $m$ it can be shown that all the items that result from dot movement are indeed generated. Finally, the completion (2) operation is applicable and (since $A \to D$) we have $C_r \sqsubseteq A^y$.

**Theorem 4.24** *If $w \in L(G)$ then the computation triggered by $w$ is successful.*

**Proof:** $w \in L(G)$, hence $Abs(A_s) \leadsto B$, where $B \in PT_w(1, n)$. Hence there exist $A', B'$ such that $A' \sqcup Abs(A_s) \neq \top, B' \sqsupseteq B$ and $A' \xrightarrow{*} B'$. By the parsing invariant, there exists $l$ such that $[0, C, n, \textsc{Comp}] \in I_l$ where $C \sqsubseteq A'$. Hence $C \sqcup Abs(A_s) \neq \top$, and therefore the computation is successful.

### 4.3.3 Subsumption check

To assure efficient computation and eliminate redundant items, many parsing algorithms employ a mechanism called *subsumption check* (see, e.g., [16, 18]) to filter out certain generated items. We introduce this mechanism below and show that it doesn't effect the correctness of the computation.

Define a (partial) order over items: $[i_1, A_1, j_1, c_1] \preceq [i_2, A_2, j_2, c_2]$ iff $i_1 = i_2, j_1 = j_2, c_1 = c_2$ and $A_1 \sqsubseteq A_2$. Modify the ordering on sets of items as follows: $I_1 \preceq I_2$ iff for every $x_1 \in I_1$ there exists $x_2 \in I_2$ such that $x_2 \preceq x_1$. Sets of items are no longer ordered by inclusion, but rather by a weaker condition that only requires the existence of a more general item (in the higher set) for every item (in the lower set).

The subsumption filter is realized by modifying $T_{G,w}$: $x \in T_{G,w}(I)$ only if there does not exist any item $x' \in T_{G,w}(I)$ such that $x' \preceq x$. Namely, for all items that span the same substring and have the same status (Act or Comp), only the most general one is preserved across successive applications of $T_{G,w}$. Given the new ordering of sets of items, it can be shown that this modification does not harm neither monotonicity nor continuity, and hence every computation is guaranteed to reach a least fix-point. Obviously, the soundness of the computation is also maintained. More interestingly, completeness is preserved, too: recall that the parsing invariant (b) states that if $A \xrightarrow{*} A' \sqsupseteq B$ then for every $k$ some item $[i_k, C_k, j_k, \textsc{Comp}]$ is generated such that $C_k \sqsubseteq A^k$. Since the subsumption test only leaves out an item if a more general one exists, the invariant still holds and hence the correctness of the computation is guaranteed. Notice that if $L(G)$ would have been defined as the set of strings that are derivable from the start symbol itself, the subsumption check might have removed crucial items, and the computation could cease to be correct.

### 4.3.4 Termination

It is well-known (see, e.g., [13, 9]) that unification-based grammar formalisms are Turing-equivalent, and therefore decidability cannot be guaranteed in the general case. However, for grammars that satisfy a certain restriction, termination of the computation can be proven. We make use of the well-foundedness result to prove that parsing is terminating for *off-line parsable* grammars.

*Off-line parsability* was introduced by [10] and adopted by [13], according to which "A grammar is off-line parsable if its context-free skeleton is not infinitely ambiguous". As [9] points out, this restriction (defined in slightly different terms) "ensures that the number of constituent structures that have a given string as their yield is bounded by a computable function of the length of that string". The problem with this definition is demonstrated by [7]: "Not every natural unification grammar has a context-free backbone". To overcome this problem, [7] uses a different restriction: "A grammar is depth-bounded if for every $L > 0$ there is a $D > 0$ such that every parse tree for a sentential form of $L$ symbols has depth less than $D$". [16] generalizes it and we use an adaptation of his definition below.

**Definition 4.25 (Finite-range generalizing functions)** *A function $F : D \to D$, where $D$ is a partially-ordered domain, is **finite-range generalizing** (FRG) iff the range of $F$ is finite and for every $d \in D, F(d) \preceq d$.*

**Definition 4.26 (Strong off-line parsability)** *A grammar G is **strongly off-line parsable** iff there exists an FRG-function F from AMRSs to AMRSs such that for every string w and AMRSs* $A, B$*, if* $A \xrightarrow{*} A' \in PT_w(i+1, j)$ *and* $B \xrightarrow{*} B' \in PT_w(i+1, j)$ *then* $F(A) \neq F(B)$.

As [16] points out, "there are non-off-line-parsable grammars for which termination holds". We use below a more general notion of this restriction: we require that $F$ distinguish $A$ and $B$ only if they are incomparable with respect to subsumption.

**Definition 4.27 (Weak off-line parsability)** *A grammar G is **weakly off-line parsable** iff there exists an FRG-function F from MRSs to MRSs such that for every string w and MRSs* $A, B$*, if* $A \xrightarrow{*} A' \in PT_w(i+1, j)$, $B \xrightarrow{*} B' \in PT_w(i+1, j)$, $A \not\sqsubseteq B$ *and* $B \not\sqsubseteq A$*, then* $F(A) \neq F(B)$.

**Theorem 4.28** *If G is weakly off-line parsable and AMRSs are acyclic then every computation terminates.*

**Proof:** Fix a computation triggered by $w$ of length $n$. We claim that there is only a finite number of generable items. Observe that the indices that determine the span of items are bounded: $0 \leq i \leq j \leq n$. It remains to show that only a finite number of AMRSs are generated. Let $x = [i, A, j, c]$ be a generated item. Suppose another item is generated where only the AMRS is different: $x' = [i, B, j, c]$ and $A \neq B$. If $A \sqsubseteq B$, $x'$ will not be preserved because of the subsumption test. There is only a finite number of AMRSs $A'$ such that $A' \sqsubseteq A$ (since subsumption is well-founded for acyclic AMRSs). Now suppose $A \not\sqsubseteq B$ and $B \not\sqsubseteq A$. By the parsing invariant (a) there exist $A', B'$ such that $A \xrightarrow{*} A' \in PT_w(i+1, j)$ and $B \xrightarrow{*} B' \in PT_w(i+1, j)$. Since $G$ is off-line parsable, $F(A) \neq F(B)$. Since the range of $F$ is finite, there are only finitely many items with equal span that are pairwise incomparable. Since only a finite number of items can be generated and the computation uses a finite number of operations, the least fix-point is reached within a finite number of steps.

The proof relies on the well-foundedness of subsumption, and indeed the proposition doesn't hold for cyclic TFSs. If cyclic TFSs are allowed, the more strict notion of strong off-line parsability is needed. Under the strong condition the above proof is applicable for the case of non-well-founded subsumption as well.

To exemplify the difference between strong and weak off-line parsability, consider a grammar $G$ that contains the following rule $R$:

$$\begin{bmatrix} \mathbf{t} & \\ f & \boxed{1} \end{bmatrix} \Rightarrow \boxed{1} \begin{bmatrix} \mathbf{t} & \\ f & \bot \end{bmatrix}$$

Assume that $\begin{bmatrix} \mathbf{t} & \\ f & \bot \end{bmatrix} \xrightarrow{*} B \in PT_w(i, j)$. This TFS is unifiable with the body (LHS) of $R$, and therefore the rule is applicable. By one application of $R$ one gets $\begin{bmatrix} \mathbf{t} & \\ f & \begin{bmatrix} \mathbf{t} & \\ f & \bot \end{bmatrix} \end{bmatrix} \xrightarrow{*} B \in PT_w(i, j)$.
This new TFS is also unifiable with the body of $R$, and thus an infinite number of TFSs can be shown to derive $B$. $G$, therefore, is not strongly off-line parsable. It is, however, weakly off-line parsable, since all the TFS that are created by successive applications of $R$ form a subsumption chain. Thus termination is guaranteed for $G$ by theorem 4.28.

# 5 Conclusions

We have formalized in this paper the concepts of *grammars* and *languages* for linguistic formalisms that are based on typed feature structures, using the notion of *multi-rooted structures* that generalize feature structures. We use multi-rooted structures for representing grammar *rules* as well as (the equivalent of) sentential forms that are generated during parsing. We described a computational process that corresponds to parsing with respect to such formalisms. We thus achieved two different specifications, namely a declarative (derivation-based) one and an algebraic (computation-based) one, for the semantics of those formalisms. Next, we have proved that the two specifications coincide, namely that the computational process induced by the algebraic specification is correct with respect to the declarative specification. Finally, we formally characterized a subset of the grammars, *off-line parsable* ones, for which termination of parsing can be guaranteed. Making use of the well-foundedness of the subsumption relation, we proved that for every grammar in this class, parsing is finitely terminating.

# Acknowledgments

This work is supported by a grant from the Israeli Ministry of Science: "Programming Languages Induced Computational Linguistics". The work of the second author was also partially supported by the Fund for the Promotion of Research in the Technion. We wish to thank Bob Carpenter and Dale Gerdemann for fruitful discussions.

# A Examples

An example of a TFS-based grammar is given in figures 6 and 7. The signature, containing the type hierarchy, the appropriateness specification and the start symbol are depicted in figure 6, whereas the rules and the lexicon are displayed in figure 7.

Type hierarchy:

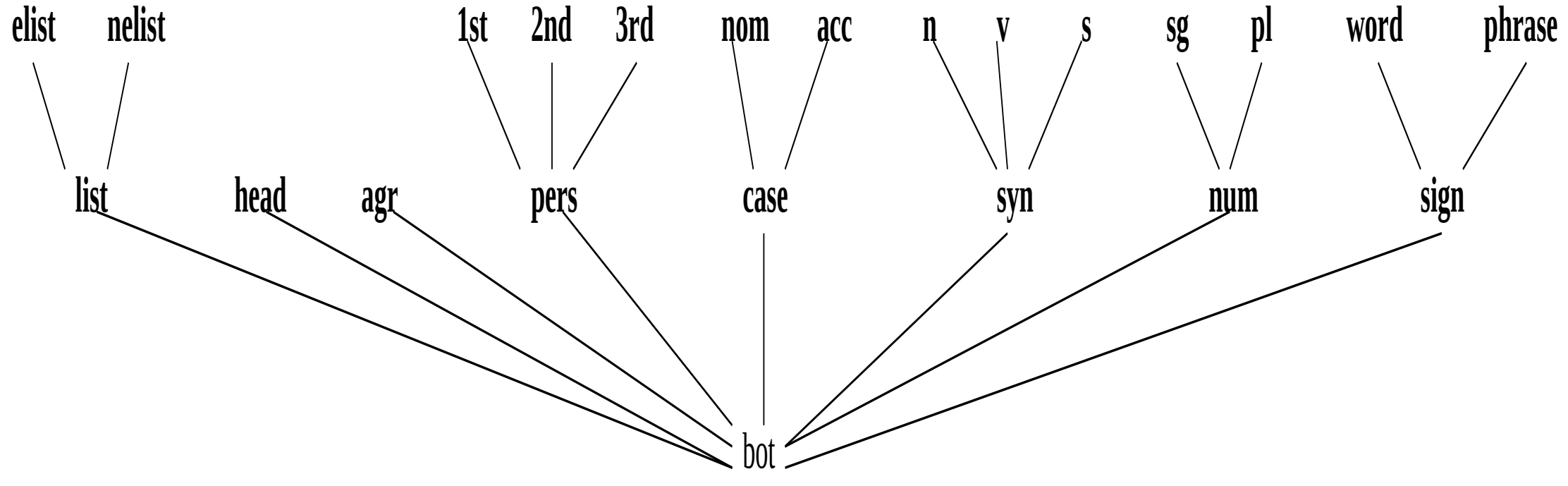

Appropriateness:

$$\begin{bmatrix} \textbf{sign} & \\ SYN: & [\textbf{syn}] \\ HEAD: & [\textbf{head}] \\ CASE: & [\textbf{case}] \\ SUBJ: & [\textbf{head}] \\ SBCT: & [\textbf{list}] \end{bmatrix} \begin{bmatrix} \textbf{nelist} & \\ 1ST: & [\textbf{sign}] \\ RST: & [\textbf{list}] \end{bmatrix} \begin{bmatrix} \textbf{head} & \\ AGR: & [\textbf{agr}] \end{bmatrix} \begin{bmatrix} \textbf{agr} & \\ PERS: & [\textbf{pers}] \\ NUM: & [\textbf{num}] \end{bmatrix}$$

Figure 6: An example type hierarchy

We simulate the process of parsing with the example grammar of figure 7 and the input "John loves fish".

Some items are added by every application of $T_{G,w}$: the items $[i, \lambda, i, \textsc{Act}]$ for $0 \leq i \leq 3$ are added due to the prediction (3) operation; the following three items are added due to the

Initial symbol:

$$\begin{bmatrix}\textbf{phrase} & \\ SYN: & [\textbf{s}]\end{bmatrix}$$

Rules:

$$\begin{bmatrix}\textbf{phrase} & \\ SYN: & [\textbf{n}] \\ HEAD: & \boxed{1}\begin{bmatrix}\textbf{head} & \\ AGR: & \boxed{3}\end{bmatrix} \\ CASE: & [\textbf{nom}]\end{bmatrix} \begin{bmatrix}\textbf{phrase} & \\ SYN: & [\textbf{v}] \\ HEAD: & \boxed{2}\begin{bmatrix}\textbf{head} & \\ AGR: & \boxed{3}\end{bmatrix} \\ SBCT: & [\textbf{elist}]\end{bmatrix} \Rightarrow \begin{bmatrix}\textbf{phrase} & \\ SYN: & [\textbf{s}] \\ SUBJ: & \boxed{1} \\ HEAD: & \boxed{2}\end{bmatrix} \tag{1}$$

$$\begin{bmatrix}\textbf{word} & \\ SYN: & [\textbf{n}] \\ HEAD: & \boxed{1} \\ CASE: & \boxed{2}\end{bmatrix} \Rightarrow \begin{bmatrix}\textbf{phrase} & \\ SYN: & [\textbf{n}] \\ HEAD: & \boxed{1} \\ CASE: & \boxed{2}\end{bmatrix} \tag{2}$$

$$\begin{bmatrix}\textbf{word} & \\ SYN: & [\textbf{v}] \\ HEAD: & \boxed{1} \\ SBCT: & \begin{bmatrix}\textbf{nelist} & \\ 1ST: & \boxed{3} \\ RST: & \boxed{2}\end{bmatrix}\end{bmatrix} \boxed{3}\begin{bmatrix}\textbf{phrase} & \\ SYN: & [\textbf{n}] \\ HEAD: & [\textbf{head}] \\ CASE: & [\textbf{acc}]\end{bmatrix} \Rightarrow \begin{bmatrix}\textbf{phrase} & \\ SYN: & [\textbf{v}] \\ HEAD: & \boxed{1} \\ SBCT: & \boxed{2}\end{bmatrix} \tag{3}$$

Lexicon:

$$John \mapsto \begin{bmatrix}\textbf{word} & \\ SYN: & [\textbf{n}] \\ HEAD: & \begin{bmatrix}\textbf{head} & \\ AGR: & \begin{bmatrix}\textbf{agr} & \\ PERS: & [\textbf{3rd}] \\ NUM: & [\textbf{sg}]\end{bmatrix}\end{bmatrix} \\ CASE: & [\textbf{case}]\end{bmatrix} \tag{4}$$

$$loves \mapsto \begin{bmatrix}\textbf{word} & \\ SYN: & [\textbf{v}] \\ HEAD: & \begin{bmatrix}\textbf{head} & \\ AGR: & \begin{bmatrix}\textbf{agr} & \\ NUM: & [\textbf{sg}]\end{bmatrix}\end{bmatrix} \\ SBCT: & \begin{bmatrix}\textbf{nelist} & \\ 1ST: & \begin{bmatrix}\textbf{phrase} & \\ SYN: & [\textbf{n}]\end{bmatrix} \\ RST: & [\textbf{elist}]\end{bmatrix}\end{bmatrix} \tag{5}$$

$$fish \mapsto \begin{bmatrix}\textbf{word} & \\ SYN: & [\textbf{n}] \\ HEAD: & \begin{bmatrix}\textbf{head} & \\ AGR: & [\textbf{agr}]\end{bmatrix} \\ CASE: & [\textbf{case}]\end{bmatrix} \tag{6}$$

Figure 7: An example grammar

scanning (5) operation:

$$[0, \begin{bmatrix}\textbf{word} & \\ SYN: & [\textbf{n}] \\ HEAD: & \begin{bmatrix}\textbf{head} & \\ AGR: & \begin{bmatrix}\textbf{agr} & \\ PERS: & [\textbf{3rd}] \\ NUM: & [\textbf{sg}]\end{bmatrix}\end{bmatrix} \\ CASE: & [\textbf{case}]\end{bmatrix}, 1, \textsc{Comp}] \tag{7}$$

$$[1, \begin{bmatrix}\textbf{word} & \\ \mathit{SYN}: & [\textbf{v}] \\ \mathit{HEAD}: & \begin{bmatrix}\textbf{head} & \\ \mathit{AGR}: & \begin{bmatrix}\textbf{agr} & \\ \mathit{NUM}: & [\textbf{sg}]\end{bmatrix}\end{bmatrix} \\ \mathit{SBCT}: & \begin{bmatrix}\textbf{nelist} & \\ 1\mathit{ST}: & \begin{bmatrix}\textbf{phrase} & \\ \mathit{SYN}: & [\textbf{n}]\end{bmatrix} \\ \mathit{RST}: & [\textbf{elist}]\end{bmatrix}\end{bmatrix}, 2, \textsc{Comp}] \tag{8}$$

$$[2, \begin{bmatrix}\textbf{word} & \\ \mathit{SYN}: & [\textbf{n}] \\ \mathit{HEAD}: & \begin{bmatrix}\textbf{head} & \\ \mathit{AGR}: & [\textbf{agr}]\end{bmatrix} \\ \mathit{CASE}: & [\textbf{case}]\end{bmatrix}, 3, \textsc{Comp}] \tag{9}$$

If $I_0 = \phi$ then $I_1$ contains only those items that are always added. $I_2 = T_{G,w}(I_1)$ and hence dot movement (1) can be applied to the prediction items and the lexical items. By unifying the first element of rule 1 with the lexical entries of both "John" and "fish", two items are generated:

$$[0, \begin{bmatrix}\textbf{phrase} & \\ \mathit{SYN}: & [\textbf{n}] \\ \mathit{HEAD}: & \begin{bmatrix}\textbf{head} & \\ \mathit{AGR}: & \begin{bmatrix}\textbf{agr} & \\ \mathit{PERS}: & [\textbf{3rd}] \\ \mathit{NUM}: & [\textbf{sg}]\end{bmatrix}\end{bmatrix} \\ \mathit{CASE}: & [\textbf{nom}]\end{bmatrix}, 1, \textsc{Act}] \tag{10}$$

$$[2, \begin{bmatrix}\textbf{phrase} & \\ \mathit{SYN}: & [\textbf{n}] \\ \mathit{HEAD}: & \begin{bmatrix}\textbf{head} & \\ \mathit{AGR}: & [\textbf{agr}]\end{bmatrix} \\ \mathit{CASE}: & [\textbf{nom}]\end{bmatrix}, 3, \textsc{Act}] \tag{11}$$

Unification of the lexical entry of "loves" with rule 3 generates the following item:

$$[1, \begin{bmatrix}\textbf{word} & \\ \mathit{SYN}: & [\textbf{v}] \\ \mathit{HEAD}: & \begin{bmatrix}\textbf{head} & \\ \mathit{AGR}: & \begin{bmatrix}\textbf{agr} & \\ \mathit{NUM}: & [\textbf{sg}]\end{bmatrix}\end{bmatrix} \\ \mathit{SBCT}: & \begin{bmatrix}\textbf{nelist} & \\ 1\mathit{ST}: & \begin{bmatrix}\textbf{phrase} & \\ \mathit{SYN}: & [\textbf{n}]\end{bmatrix} \\ \mathit{RST}: & [\textbf{elist}]\end{bmatrix}\end{bmatrix}, 2, \textsc{Act}] \tag{12}$$

Both lexical items for "John" and for "fish" can unify with the body of rule 2 and thus the completion (2) operation is applicable, and the following two items are generated:

$$[0, \begin{bmatrix}\textbf{phrase} & \\ \mathit{SYN}: & [\textbf{n}] \\ \mathit{HEAD}: & \begin{bmatrix}\textbf{head} & \\ \mathit{AGR}: & \begin{bmatrix}\textbf{agr} & \\ \mathit{PERS}: & [\textbf{3rd}] \\ \mathit{NUM}: & [\textbf{sg}]\end{bmatrix}\end{bmatrix} \\ \mathit{CASE}: & [\textbf{case}]\end{bmatrix}, 1, \textsc{Comp}] \tag{13}$$

$$[2, \begin{bmatrix} \textbf{phrase} & \\ \mathit{SYN}: & [\textbf{n}] \\ \mathit{HEAD}: & \begin{bmatrix} \textbf{head} & \\ \mathit{AGR}: & [\textbf{agr}] \end{bmatrix} \\ \mathit{CASE}: & [\textbf{case}] \end{bmatrix}, 3, \textsc{Comp}] \qquad (14)$$

Due to this last item, the next application of $T_{G,w}$ adds to $I_3$ the following item:

$$[1, \begin{bmatrix} \textbf{phrase} & \\ \mathit{SYN}: & [\textbf{v}] \\ \mathit{HEAD}: & \begin{bmatrix} \textbf{head} & \\ \mathit{AGR}: & \begin{bmatrix} \textbf{agr} & \\ \mathit{NUM}: & [\textbf{sg}] \end{bmatrix} \end{bmatrix} \\ \mathit{SBCT}: & \begin{bmatrix} \textbf{nelist} & \\ \mathit{1ST}: & \boxed{3} \\ \mathit{RST}: & [\textbf{elist}] \end{bmatrix} \end{bmatrix} \boxed{3} \begin{bmatrix} \textbf{phrase} & \\ \mathit{SYN}: & [\textbf{n}] \\ \mathit{HEAD}: & \begin{bmatrix} \textbf{head} & \\ \mathit{AGR}: & [\textbf{agr}] \end{bmatrix} \\ \mathit{CASE}: & [\textbf{acc}] \end{bmatrix}, 3, \textsc{Act}] \qquad (15)$$

and since it unifies with the body of rule 3, completion can be applied and the following item is obtained in $I_4$:

$$[1, \begin{bmatrix} \textbf{phrase} & \\ \mathit{SYN}: & [\textbf{v}] \\ \mathit{HEAD}: & \boxed{1} \begin{bmatrix} \textbf{head} & \\ \mathit{AGR}: & \begin{bmatrix} \textbf{agr} & \\ \mathit{NUM}: & [\textbf{sg}] \end{bmatrix} \end{bmatrix} \\ \mathit{SBCT}: & [\textbf{elist}] \end{bmatrix}, 3, \textsc{Comp}] \qquad (16)$$

This item combines with item 13 to create in $I_5$, by applying dot movement according to rule 1, the item:

$$[0, \begin{bmatrix} \textbf{phrase} & \\ \mathit{SYN}: & [\textbf{n}] \\ \mathit{HEAD}: & \begin{bmatrix} \textbf{head} & \\ \mathit{AGR}: & \boxed{3} \begin{bmatrix} \textbf{agr} & \\ \mathit{PERS}: & [\textbf{3rd}] \\ \mathit{NUM}: & [\textbf{sg}] \end{bmatrix} \end{bmatrix} \\ \mathit{CASE}: & [\textbf{nom}] \end{bmatrix} \begin{bmatrix} \textbf{phrase} & \\ \mathit{SYN}: & [\textbf{v}] \\ \mathit{HEAD}: & \begin{bmatrix} \textbf{head} & \\ \mathit{AGR}: & \boxed{3} \end{bmatrix} \\ \mathit{SBCT}: & [\textbf{elist}] \end{bmatrix}, 3, \textsc{Act}] \qquad (17)$$

which unifies with the body of rule 1 so that completion can generate

$$[0, \begin{bmatrix} \textbf{phrase} & \\ \mathit{SYN}: & [\textbf{s}] \\ \mathit{SUBJ}: & \begin{bmatrix} \textbf{head} & \\ \mathit{AGR}: & \boxed{3} \begin{bmatrix} \textbf{agr} & \\ \mathit{PERS}: & [\textbf{3rd}] \\ \mathit{NUM}: & [\textbf{sg}] \end{bmatrix} \end{bmatrix} \\ \mathit{HEAD}: & \begin{bmatrix} \textbf{head} & \\ \mathit{AGR}: & \boxed{3} \end{bmatrix} \end{bmatrix}, 3, \textsc{Comp}] \qquad (18)$$

This final item is more specific than the initial symbol of the grammar; it spans the entire input; hence the input "John loves fish" is a sentence of the grammar.